\documentclass{article}

\usepackage[main, final]{neurips_2026}

\usepackage[utf8]{inputenc} 
\usepackage[T1]{fontenc}    
\usepackage{hyperref}       
\usepackage{url}            
\usepackage{booktabs}       
\usepackage{amsfonts}       
\usepackage{nicefrac}       
\usepackage{microtype}      
\usepackage{xcolor}         

\usepackage{amsmath}
\usepackage{amssymb}
\usepackage{mathtools}
\usepackage{amsthm}

\usepackage[capitalize,noabbrev]{cleveref}

\theoremstyle{plain}
\newtheorem{theorem}{Theorem}[section]
\newtheorem{proposition}[theorem]{Proposition}
\newtheorem{lemma}[theorem]{Lemma}

\theoremstyle{definition}
\newtheorem{definition}[theorem]{Definition}
\newtheorem{assumption}[theorem]{Assumption}
\theoremstyle{remark}

\usepackage[textsize=tiny]{todonotes}

\usepackage{float}

\usepackage{amssymb}
\usepackage{amsmath}
\usepackage{amsthm}

\newcommand{\HV}{\text{HV}}

\usepackage{graphicx}
\usepackage{subcaption}
\usepackage{caption}
\usepackage{float}

\usepackage{hyperref}

\usepackage{algorithm}
\usepackage{algorithmic}

\usepackage{tabularx}  
\usepackage{booktabs}  
\usepackage{pifont}    
\usepackage{amssymb}   
\usepackage{ragged2e}  
\usepackage{tikz}
\usepackage{pgfplots}
\pgfplotsset{compat=1.18}

\newcolumntype{C}[1]{>{\centering\arraybackslash}m{#1}}
\newcolumntype{Y}{>{\centering\arraybackslash}X}

\title{MOBO-CAPS: Multi-Objective Bayesian Optimization \\ with Cardinality-Aware Pareto Selection}

\author{%
  Hanyang Wang \\
  University of Warwick\\
  Coventry, CV4 7AL, United Kingdom \\
  \texttt{Hanyang.Wang@warwick.ac.uk} \\
  \and
  \textbf{Juergen Branke} \\
  University of Warwick\\
  Coventry, CV4 7AL, United Kingdom \\
  \texttt{Juergen.Branke@wbs.ac.uk} \\
  \and
\textbf{Matthias Poloczek}\thanks{This research does not relate to Matthias’ work at Amazon and Databricks.} \\
Databricks \\
San Francisco, CA 94105, USA \\
\texttt{matthias.poloczek@databricks.com} \\
}

\begin{document}

\maketitle

\begin{abstract}
Multi-objective Bayesian optimization (MOBO) traditionally aims to approximate the entire Pareto front, but, due to resource constraints, real-world deployment is often limited to testing a small, representative subset of solutions. 
Hence current methods first approximate the full front before selecting solutions, potentially wasting costly evaluations. We formalize this problem as Multi-objective Bayesian Optimization with Cardinality-aware Pareto Selection (MOBO-CAPS), where the goal is to identify a set of exactly $M$ solutions with high hypervolume. We propose two practical cardinality-aware acquisition strategies: REHVI, a replacement-aware approximation to expected hypervolume improvement, and CHO-UCB, which constructs an optimistic $M$-point set by optimizing a collective UCB-based hypervolume objective. 
We prove a sublinear simple regret bound for the batched version of CHO-UCB, and empirically show that cardinality-aware acquisition design improves over standard MOBO followed by post-hoc subset selection across nine benchmarks.
\end{abstract}

\section{Introduction}
Bayesian Optimization (BO) has become a standard method for the optimization of expensive-to-evaluate black-box functions, a task that is ubiquitous in science and engineering. Multi-objective Bayesian Optimization (MOBO) aims to identify the Pareto front, i.e., a set of non-dominated solutions representing different trade-offs between conflicting objectives. A common approach in MOBO is to sequentially select evaluation points that improve coverage of the Pareto front, with hypervolume being a widely used performance metric due to its Pareto compliance and ability to capture both convergence and diversity. This approach has achieved considerable success in practical applications, such as protein design \citep{stanton2022accelerating}, materials engineering \citep{biswas2021multi} and model predictive control \citep{makrygiorgos2022performance}. 

However, in real-world applications, decision-makers often require only a small, carefully selected subset of Pareto-optimal solutions rather than the entire front. The motivations for considering only a small, finite subset of solutions are twofold. First, the goal may be to find a small but representative set that covers the entire Pareto front. This is common when catering to a heterogeneous user base with diverse preferences, such as providing a range of investment options with different risk profiles or car models that appeal to different market segments. The aim is to present a wide spectrum of possible trade-offs. Second, MOBO is often only the first step in a longer decision-making process. Engineers and scientists frequently require a manageable candidate set for detailed, hands-on evaluation. This subsequent scrutiny involves domain-specific criteria that are difficult or impossible to formalize as objectives, such as model interpretability, aesthetic appeal of a design, or subtle safety considerations. This detailed review might itself focus on a diverse set of candidates from across the Pareto front, or it may concentrate on a specific sub-region of the Pareto front deemed most promising. In practice, this leads to a two-step procedure: run MOBO to approximate the entire Pareto front, then apply a post-hoc selection procedure to extract a subset of $M$ solutions. One way to select a subset of solutions is to pick the $M$ solutions that jointly maximize the hypervolume, i.e., the space dominated by the solutions, bounded by some given reference point. This problem itself is NP-hard \citep{chen2021fast}, and various heuristics have been proposed to solve it, see, e.g., \citet{bringmann2010efficient, kuhn2016hypervolume}. Despite its convenience, this strategy remains informal, and the cardinality constraint is not incorporated into the optimization process itself, making it potentially inefficient.

In this paper, we formalize this overlooked, yet fundamental setting as a sequential black-box optimization problem: identifying exactly $M$ solutions with maximal hypervolume under a limited evaluation budget. We refer to this problem as \emph{Multi-objective Bayesian Optimization with Cardinality-Aware Pareto Selection} (MOBO-CAPS). 
This formulation bridges the gap between theoretically founded MOBO algorithms and the practical need to select a small number of actionable solutions. 
It also opens up new methodological questions, such as how to design acquisition functions that effectively optimize the hypervolume under a fixed-cardinality constraint. 
Hypervolume serves as our objective for selecting $M$ solutions because it guarantees that the chosen solutions lie on the Pareto front, while promoting coverage of diverse trade-offs among objectives. By targeting the $M$-point set that maximizes the hypervolume, we preserve alignment with the fundamental principles of MOBO. 

To tackle this problem, we propose two algorithms. The first, \textit{Replacement-aware Expected Hypervolume Improvement} (REHVI), extends the classic Expected Hypervolume Improvement (EHVI) acquisition function by accounting for the need to maintain a solution set of size $M$. The second, \textit{Collective Higher-dimensional Optimization with Upper Confidence Bound} (CHO-UCB), reformulates the problem as a joint optimization over $M$ solutions using a high-dimensional acquisition function. We analyze the \textcolor{black}{computational} efficiency and optimization behavior of these methods, and conduct extensive experiments across both synthetic benchmarks and real-world applications. Our results show that incorporating the cardinality constraint throughout the optimization process leads to better performance than traditional MOBO followed by post-hoc selection. 

Our contributions are fourfold. First, we introduce MOBO-CAPS, a formal MOBO setting in which the final recommendation is constrained to contain exactly $M$ solutions. Second, we propose two new algorithms, REHVI and CHO-UCB, that directly incorporate cardinality information into the optimization process. 
Third, we establish a regret bound for batched CHO-UCB, providing a first regret analysis for this setting. Finally, we validate our approach through extensive experiments on synthetic and real-world benchmarks, demonstrating that our methods outperform standard MOBO with post-hoc selection.

The structure of this paper is as follows. Section 2 reviews the literature on MOBO and related problems. Section 3 introduces the MOBO-CAPS problem setting. Section 4 presents the REHVI and CHO-UCB methods. Section 5 reports experimental results. Section 6 concludes with a summary and discusses avenues for future work. 
The code is available at \href{https://github.com/HanyangHenry-Wang/MOBO-CAPS/tree/master}{https://github.com/HanyangHenry-Wang/MOBO-CAPS/tree/master}.

\section{Literature Review}
In multi-objective optimization, various metrics have been proposed to measure the quality of a Pareto front approximation. The most prominent, and the one that is strictly Pareto compliant, is hypervolume \citep{zitzler1998multiobjective}. 
In MOBO, the EHVI acquisition function directly optimizes hypervolume \citep{emmerich2007gradient}, though computational costs initially limited adoption. \citet{yang2019efficient} improved \textcolor{black}{computational efficiency} by decomposing integration domains into axis-aligned hyperrectangles, while \citet{yang2019multi} introduced gradient-based optimization for bi-objective problems. \citet{daulton2020differentiable} leveraged automatic differentiation and parallel hardware to enhance scalability significantly. Alternative hypervolume-based approaches include UCB-based hypervolume improvement estimation called $\mathcal{S}$-Metric-Selection-based Efficient Global Optimization (SMS-EGO) \citep{ponweiser2008multiobjective} and probability of hypervolume improvement \citep{couckuyt2014fast}.
Beyond methods based on hypervolume, ParEGO \citep{knowles2006parego} converts multi-objective problems into a sequence of single-objective problems by applying a different weighted Tchebycheff scalarization at each iteration. A related approach, proposed by \citet{paria2020flexible}, maintains separate surrogate models for each objective and applies the scalarization directly within the acquisition function.

Four related multi-objective optimization problem settings focus on selecting single points or subsets from Pareto fronts. Bayesian Optimization with Preference Learning \citep{astudillo2020multi,lin2022preference,ungredda2022single,ozaki2024multi,wang2025bayesian} selects solutions maximizing decision maker utility based on learned preferences. Finding Knees in Bayesian Multi-objective Optimization \citep{heidari2022finding,heidari2024knee}, aims to identify a single Pareto optimal point maximizing hypervolume. 
However, searching for a single solution with maximal hypervolume is much easier, as it avoids the combinatorial problem of identifying a subset. The third similar problem setting is Multi-Objective Coverage Bayesian Optimization \citep{maus2025covering}. It selects solution subsets in many-objective problems, maximizing the so-called coverage score, defined as the sum over all objective functions of the maximum observed across all solutions in the set of given cardinality (with the cardinality being less than the number of objectives). Finally, while not a BO method, \citet{wang2017hypervolume} addressed a closely related problem for functions with accessible gradient. \textcolor{black}{Their method performs gradient ascent directly in the decision space to discover a set of Pareto-optimal solutions.} A direct comparison of these settings is provided in Appendix \ref{literature_table}.

For a given Pareto front, researchers have looked at the problem of identifying a small number of solutions with maximal hypervolume. The $\mu$-distribution problem identifies $\mu$ points from a known Pareto front to maximize hypervolume. For example, \citet{glasmachers2014optimized} developed a gradient ascent approach \textcolor{black}{that operates directly in the objective space to compute hypervolume‐optimal $\mu$-distributions for two- and three-objective benchmark fronts.} Theoretical studies have also examined optimal $\mu$-distributions, e.g., \citet{emmerich2007gradient,auger2009theory} showed that hypervolume-optimal points are uniformly distributed on linear Pareto fronts in two-objective problems.  The Hypervolume Subset Selection Problem (HSSP) selects a fixed number of solutions from a given pool to maximize hypervolume. \citet{ishibuchi2009selecting} proposed genetic algorithms encoding solutions as binary strings with fitness combining hypervolume and a penalty term. Later work \citep{ishibuchi2014selecting} reformulated HSSP as multi-objective optimization with different reference points. Several evolutionary algorithms have been proposed for $\mu$-distribution and HSSP; see \citet{shang2020survey} for comprehensive coverage. \textcolor{black}{Unlike these settings, MOBO-CAPS is a sequential black-box problem in which the optimal set for a given size~$M$ must be composed rather than selected from a pool of already observed sets.}

The task of finding a set of solutions in single-objective BO has also been studied. \citet{maus2023discovering} select diverse solutions according to given criteria by maintaining several trust regions.  \citet{mei2023bayesian} identify local optima through BO by incorporating the surrogate model's first-order derivatives. \citet{miller2024expected} proposed Expected Diverse Utility, which identifies diverse $\epsilon$-optimal solutions (locally optimal within tolerance $\epsilon$ of the global optimum) using a closed-form acquisition function that enables efficient optimization. \citet{kent2024bayesian} extended BO to quality-diversity search by modeling both objectives and descriptors with Gaussian processes (GPs) and proposing an Expected Improvement (EI)-based acquisition to efficiently identify the best solution in each predefined region of the descriptor space.

\section{Problem Statement}
In this section, we formally introduce the Multi-objective Bayesian Optimization
with Cardinality-Aware Pareto Selection (MOBO-CAPS) problem.

The goal is to maximize $K$ \textcolor{black}{expensive} objective functions $\mathbf{f} = (f_1,\ldots,f_K)$ over the search space $\mathcal{X} \subseteq \mathbb{R}^d$.
In this paper, we use \emph{solution} to refer to an input $x\in \mathcal{X}$ and \emph{point} to refer to an output $y\in \mathbb{R}^K$. 
That is, a point $\mathbf{f}(x)$ is given by the objective values obtained by the solution~$x$, observed without noise.
We use the standard maximization convention: a solution \(x\) is Pareto optimal if no \(x'\in\mathcal X\) satisfies \(\mathbf f(x')\ge \mathbf f(x)\) componentwise with strict inequality in at least one objective. 

We measure the quality of a finite set of objective vectors via the \emph{hypervolume} (HV) indicator.
Given a reference point $z_{\mathrm{ref}}\in\mathbb{R}^K$ dominated by all points of interest, the hypervolume of $Y\subset\mathbb{R}^K$ is the Lebesgue measure of the region dominated by $Y$ and bounded by $z_{\mathrm{ref}}$, $\HV(Y; z_{\mathrm{ref}})
:= \lambda\!\left(\bigcup_{y\in Y}\,[z_{\mathrm{ref},1},y_1]\times\cdots\times[z_{\mathrm{ref},K},y_K]\right)$, where $\lambda(\cdot)$ denotes the $K$-dimensional Lebesgue measure. We write $\HV(Y)$ when $z_{\mathrm{ref}}$ is clear from context.

We use \emph{CAPS-set} to refer to a solution set of cardinality $M$ returned by an algorithm as a cardinality-constrained approximation of the Pareto set.
Our goal is to find a CAPS-set ${X}^\star:=\{x^\star_1,\ldots,x_M^\star\}$ that maximizes the hypervolume:
\[
X^\star \in \arg\max_{\substack{X\subseteq \mathcal{X} \\ |X|=M}} \HV\!\bigl(\mathbf{f}(X)\bigr),
\]
where ${X} = \{x_1,\ldots,x_M\}$ denotes a set of $M$ solutions and $\mathbf{f}(X):=\{\mathbf{f}(x):x\in X\}$.

In what follows, $\mathcal{P}_t$ denotes the current approximate Pareto set in iteration~$t$, and $\mathcal{P}^M_t$ is the currently best CAPS-set in iteration~$t$.
Moreover, $\mathbf{f}(x)$ gives the true objective values of $x$, while \(\hat{\mathbf f}(x)\) denotes the posterior predictive distribution induced by the \(K\) surrogate models.

\textcolor{black}{Notice that as $M$ becomes large, the MOBO-CAPS problem gradually approaches the standard MOBO setting, since returning many Pareto-optimal solutions increasingly resembles approximating the full Pareto front. In this regime, existing MOBO methods are often sufficient, especially when combined with a post-hoc selection procedure. Conversely, the cardinality constraint is most meaningful when only a small number of final solutions is required. A small value of the cardinality parameter $M$ is common in real-world applications. For example, a car company may only release three vehicle variants, such as an economy model, a balanced model, and a high-performance model. Similarly, in pharmaceutical product development, a company may optimize a topical pain-relief gel for absorption, irritation risk, and manufacturing cost, but only commercialize two strengths, such as regular-strength and higher-strength formulations, due to regulatory and production requirements. These practical constraints motivate our focus on small values of $M$ in the experiments. This choice is also motivated by the empirical results in Appendix~\ref{Full Experimental Results}, where the performance gap between MOBO-CAPS and standard MOBO with post-hoc selection generally becomes smaller as $M$ increases, although this trend is not consistent across all test problems.}

\section{Proposed Approach}
A naive approach for MOBO-CAPS is post-hoc selection, where we run a standard MOBO algorithm to approximate the Pareto front and then extract the $M$ solutions that maximize the hypervolume. 
This approach is conceptually simple and theoretically sound. If MOBO recovers the true Pareto set, post-hoc selection will identify the optimal CAPS-set. However, this approach raises \textcolor{black}{sample} efficiency concerns, as it utilizes cardinality information only at the final step rather than incorporating the size constraint already during optimization. 
Figures~\ref{RHVI_plots} and~\ref{af_vis} show examples where a MOBO algorithm will evaluate solutions that contribute to the total hypervolume but are not useful for maximizing the hypervolume of a best CAPS-set, because in a cardinality-constrained setting a new solution has to replace another solution. 

To address these limitations by leveraging cardinality information during optimization, one could consider a Bayes-optimal one-step approach that selects the next solution maximizing the expected improvement of the CAPS-set hypervolume among all observed solutions. However, this method requires computationally expensive Monte Carlo simulation and demands identification of the optimal CAPS-set for each fantasy value, rendering it impractical for real-world applications.

Therefore, we propose two computationally tractable methods that efficiently incorporate cardinality information to improve MOBO performance: Replacement-aware Expected Hypervolume Improvement (REHVI) and Collective Higher-dimensional Optimization with Upper Confidence Bound (CHO-UCB). REHVI is a variant of EHVI that accounts for a fixed-cardinality solution set by removing one existing point from the current CAPS-set when a new point is added. On the other hand, CHO-UCB reformulates the multi-objective optimization problem as a higher-dimensional single-objective problem, enabling joint selection of multiple solutions via a collective acquisition function.

\subsection{Replacement-aware Expected Hypervolume Improvement (REHVI)}
Given the sample efficiency of EHVI in MOBO, we adapt this acquisition function to our problem setting as follows. Assuming the current best CAPS-set in iteration $t$ is $\mathcal{P}^M_t$, the replacement-aware hypervolume improvement of a new point $y$ is calculated as 
\begin{equation*}
\label{eq:RHVI}
\begin{split}
\text{RHVI}(y) =
\max_{y' \in \mathbf{f}(\mathcal{P}_t^M) \cup \{y\} } & \Big\{ \text{HV}\big({\mathbf{f}}(\mathcal{P}_t^M) \cup \{y\} \setminus \{y'\} \big) - \text{HV}\big({\mathbf{f}}(\mathcal{P}^M_t)\big)\Big\}.
\end{split}
\end{equation*}

The idea behind this formula is that we can only maintain $M$ solutions in the CAPS-set $\mathcal{P}_t^M$. Therefore, when we introduce a new point $y$, we must remove a point $y'$. Naturally, we should remove the element that results in the largest gain in hypervolume. 

Therefore, the expected improvement when evaluating a solution $x$ is 
$$\text{REHVI}(x) = \mathbb{E}[\text{RHVI}({\hat{\mathbf{f}}(x)})],$$
where the expectation is taken with respect to the posterior predictive distribution of $\hat{\mathbf{f}}$.

\begin{figure}[h]
	\centering
 \includegraphics[width=0.9\linewidth]{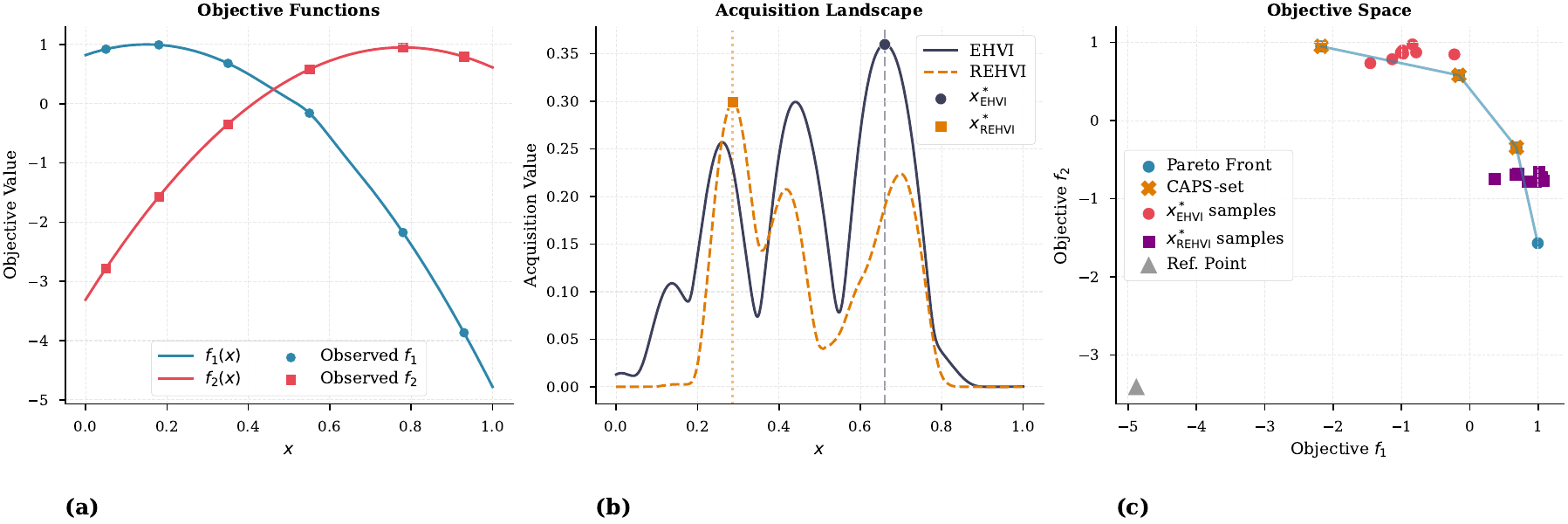}
    \vspace*{-1mm}
	\caption{Difference between REHVI and EHVI on a bi-objective problem. (a) The two underlying objective functions and the observed solutions. (b) The acquisition landscapes of REHVI and EHVI, with their respective maximizers. (c) The current CAPS-set and posterior samples of the solutions selected by EHVI and REHVI. The REHVI-selected solution (purple) yields a larger replacement-aware hypervolume gain, despite contributing less to the total hypervolume on its own.}
	\label{RHVI_plots}
        \vspace*{-2mm}
\end{figure}

The acquisition function is called Replacement-aware Expected Hypervolume Improvement (REHVI). An illustration highlighting the difference in solution selection between REHVI and EHVI is provided in Figure~\ref{RHVI_plots}. Since REHVI is defined as a cardinality-aware generalization of EHVI, when the current non-dominated set has fewer than $M$ points, replacement is not yet meaningful, and the criterion reduces to standard EHVI. Once $|\mathcal{P}_t|\ge M$, REHVI evaluates the improvement obtained by replacing one point in the current incumbent CAPS-set.

The implementation and computational complexity of the REHVI acquisition function are discussed in Appendix~\ref{appendix_time}. The full MOBO-CAPS algorithm using REHVI is given in Appendix~\ref{Pseudocode}.

\textcolor{black}{We emphasize that our proposed REHVI is distinct from NEHVI (Noisy Expected Hypervolume Improvement) introduced by \citep{daulton2021parallel}. The contributions are orthogonal: while NEHVI extends EHVI to handle uncertainty from noisy objective evaluations, REHVI modifies the EHVI formulation to account for the replacement dynamics required when optimizing under a strict, fixed-cardinality constraint.}

\subsection{Collective Higher-dimensional Optimization with Upper Confidence Bound (CHO-UCB)}
While REHVI optimizes the CAPS-set through iterative single-point replacements, 
an alternative perspective is to directly treat the entire CAPS-set as a collective decision variable. Following this idea, we propose Collective Higher-dimensional Optimization (CHO): rather than approaching this as a $d$-dimensional multi-objective optimization problem, we formulate it as a single $M\times d$-dimensional problem that optimizes all $M$ solutions jointly. 

A natural starting point for CHO is a set-valued extension of EI:
$\text{CHO-EI}(X)=\mathbb{E}\!\left[\left\{\mathrm{HV}(\hat{\mathbf f}(X))-\mathrm{HV}\!\left(\mathbf f(\mathcal P_t^M)\right)\right\}^{+}\right]$,
where \(X=\{x_1,\dots,x_M\}\) is a set of \(M\) solutions and \(\mathcal P_t^M\) is the current best CAPS-set. In practice, however, CHO-EI is difficult to optimize: as the incumbent hypervolume approaches its optimum, the acquisition landscape becomes flat, the expected improvement vanishes, and gradient-based optimization stalls, even for moderate \(M\). We therefore do not pursue CHO-EI further. Additional discussion and ablations are provided in Appendix~\ref{acquisition_value} and Appendix~\ref{raasp_ablation}.

Instead, we propose Collective Higher-dimensional Optimization with Upper Confidence Bound (CHO-UCB), which preserves the collective set-based formulation but computes the hypervolume of an optimistic $M$-point set under GP upper confidence bounds. This avoids Monte Carlo estimation and provides a smoother acquisition landscape. 

Following standard procedure, we calculate the UCB as
$\text{UCB}({X})=\mu({X})+\sqrt{\beta_t}   \sigma({X}),$
  where $\mu(\cdot)$ and $\sigma(\cdot)$ denote the posterior mean and standard deviation of the surrogate model at solutions $X$, and $\beta_t >0 $ is a hyperparameter trading off exploration and exploitation. We then approximate the Pareto front using these UCB values and compute the corresponding hypervolume: 
  $$\text{CHO-UCB}({X}) = \text{HV}(\text{UCB}({X})),$$
  where ${X} = \{x_1 , \ldots , x_M \}$ is a set of $M$ solutions.

A key advantage of the CHO-UCB formulation is that it admits a theoretical guarantee: under standard RKHS assumptions, its simple regret converges to zero asymptotically as the number of samples increases. \textcolor{black}{While regret guarantees have been established for standard MOBO, including cumulative regret bounds \citep{chowdhury2021no} and sublinear hypervolume regret achieved via scalarizations \citep{zhang2024optimal}, these foundational analyses assume an unconstrained Pareto set approximation, and they do not translate to settings with strict budget limits on the solution set size. To the best of our knowledge, our theoretical analysis provides the first simple regret bound specifically derived for MOBO under fixed-cardinality constraints.}

A key technical challenge in analyzing CHO-UCB is that the hypervolume indicator is defined over sets of points rather than individual points, which prevents the direct application of standard UCB proof techniques. We resolve this by introducing the matching distance on multisets (Definition~\ref{matching_distance}) and proving that the hypervolume indicator is Lipschitz continuous with respect to this distance (Lemma~\ref{lem:hv-lipschitz}). This result, which may be of independent interest, allows us to reduce the set-valued regret to pointwise posterior uncertainty bounds.

Our analysis requires two standard regularity conditions. The first ensures hypervolume calculations are well-defined, while the second guarantees the objective functions are compatible with Gaussian process (GP) modeling.

\begin{assumption}[Bounded objectives]
\label{ass:box_main}
We fix a reference point $z_{\mathrm{ref}} \in \mathbb{R}^K$ and an upper bound vector $u \in \mathbb{R}^K$ satisfying strict dominance $u \succ z_{\mathrm{ref}}$.
We assume that the true objective functions are bounded from above by $u$:
\[
    \mathbf{f}(x) \preceq u \quad \text{for all } x \in \mathcal{X}.
\]
\end{assumption}

\begin{assumption}[RKHS regularity]
\label{ass:rkhs-batch-simple_main}
For each $i \in [K]$, $f_i$ lies in the RKHS $\mathcal{H}_{k_i}$ of a known
positive definite kernel $k_i : \mathcal{X} \times \mathcal{X} \to \mathbb{R}$.
Moreover, $k_i(x,x) \le 1$ for all $x \in \mathcal{X}$ and
$\|f_i\|_{\mathcal{H}_{k_i}} \le B$ for some $B > 0$.
\end{assumption}

\begin{theorem}[Noise-free Simple Regret of Batched CHO-UCB]
\label{main_thm}
Suppose the objective functions $f_i$ satisfy the RKHS Assumption~\ref{ass:rkhs-batch-simple_main} with bound $B$, and the problem domain is bounded as per Assumption~\ref{ass:box_main} with $u = (2B, \dots, 2B)$.
Setting the exploration parameter $\sqrt{\beta_t} = B$ for all $t$, the simple regret of Batched CHO-UCB satisfies:
{\small
\[
  r_T^{\mathrm{simple}}
  \;\le\;
  \frac{2 B MK(2B - \min_j z_{\mathrm{ref}, j})^{K-1}}{T} \sum_{i=1}^K C_i(T, d, \theta_i),
\]
}

where $C_i(T, d, \theta_i)$ denotes the cumulative posterior standard deviation bound given in Lemma~3 of \citet{iwazaki2025gaussian}, and $\theta_i$ collects the hyperparameters of the $i$-th objective model. For ease of reference, we restate this result as Lemma~\ref{lemma:posterior_std_bound}.
\end{theorem}

Due to space constraints, the full proof for the batched CHO-UCB formulation is deferred to Appendix~\ref{full_proof}. The analysis follows an optimism-based UCB framework, together with one new geometric ingredient tailored to our set-valued hypervolume objective. First, valid confidence intervals for the objective functions imply that the instantaneous regret can be bounded by the gap between the hypervolume of the upper and lower confidence bounds of the selected set. To relate this quantity to pointwise posterior uncertainty,  Lemma~\ref{lem:hv-lipschitz} shows that the hypervolume indicator is Lipschitz continuous with respect to the matching distance on multisets (Definition~\ref{matching_distance}). This lets us upper bound the regret by the maximum of posterior standard deviations at the queried points. Finally, combining this reduction with the cumulative uncertainty bounds of \citet{iwazaki2025gaussian} yields a sublinear simple regret bound. 

The implementation and computational complexity of the CHO-UCB acquisition function are discussed in Appendix~\ref{appendix_time}. 

The CHO-UCB formulation analyzed above is batched: all \(M\) points selected by the collective UCB objective are evaluated in each round. In practice, many applications allow only sequential evaluations, so we use a scheduler-based variant: CHO-UCB first constructs the same optimistic \(M\)-point candidate set, and an EHVI scheduler selects one point for evaluation. The scheduler is a practical sequentialization layer that enables faster feedback and avoids committing to an entire batch. 
Thus, the regret theorem applies to the underlying batched rule, while analysis of the scheduler-based variant is left for future work. \textcolor{black}{Empirically (Appendix~\ref{scheduler}), the EHVI scheduler outperforms batched evaluation under equal budgets. 
We additionally compare three schedulers: random, uncertainty (variance) and SMS‑EGO. 
The random scheduler performs worst, followed by the batched variant, while the remaining three schedulers achieve comparable performance. This insensitivity to the specific scheduling heuristic indicates that the cardinality‑aware set construction is a key performance driver, whereas the selection criterion is not. 
In the experiments, "CHO‑UCB" refers to this EHVI scheduler‑based variant unless otherwise noted.}

\section{Experimental Evaluation}
\label{Experiments}
We study five commonly-used synthetic functions: BraninCurrin (2D, 2 objectives), 
DTLZ1 (3D, 2 objectives), DTLZ2 (3D, 2 objectives), VLMOP2 (6D, 2 objectives), and VLMOP3 (2D, 3 objectives). To examine the effect of input dimensionality, we additionally evaluate DTLZ2 with 6D and 10D inputs for $M = 3$. Besides, to examine the effect of the number of objectives on performance, we evaluate DTLZ2 with 6D inputs and 2, 3, and 4 objectives. We also consider four real-world tasks spanning diverse application domains, including drug synthesis, 
chemical manufacturing, and mechanical design: VehicleSafety (5D, 3 objectives; \citealp{tanabe2020easy}), 
Truss Design (4D, 2 objectives; \citealp{cheng1999generalized}), Penicillin (7D, 3 objectives; 
\citealp{liang2021scalable}), and SnAr (4D, 2 objectives; \citealp{hone2017rapid}). Detailed problem descriptions are provided in Appendix~\ref{Real-life Test Problems}. \\\\
Each problem is evaluated for \(M \in \{1,2,3,4,5\}\), yielding 45 test cases in total. We focus on relatively small $M$, since MOBO-CAPS is primarily motivated by applications in which the decision maker can only inspect or deploy a small set of solutions; as \(M\) becomes large, the problem increasingly approaches standard MOBO. In Figure~\ref{hv_percentage_bar} of Appendix~\ref{Real-life Test Problems}, we provide several examples of the hypervolume coverage of the CAPS-set relative to the full Pareto front. Across the four examples, the \(M=3\) CAPS-set covers between \(69\%\) and \(99.7\%\) of the Pareto-front hypervolume, further supporting our focus on small cardinalities. Furthermore, while practical considerations prioritize small $M$, we also evaluate $M=10,15$ on DTLZ2 to demonstrate the scalability of REHVI and CHO-UCB. \\\\
We compare against a broad suite of state-of-the-art methods that are applicable to these problems, although they have not been specifically designed for them:
Sobol, EHVI, SMS-EGO~\citep{ponweiser2008multiobjective},
ParEGO \citep{botorch_mobo_tutorial}, its reference-point variant denoted as \text{ParEGO}$^*$,
JES~\citep{tu2022joint} and HV-KG \citep{daulton2023hypervolume}. JES and HV-KG exploit knowledge of the target recommendation size by setting the number of Pareto samples or posterior-mean Pareto candidates to~\(M\). Detailed descriptions of these benchmark algorithms are provided in Appendix~\ref{Benchmark Algorithm}. Note that REHVI can be viewed as a cardinality-aware variant of EHVI, while CHO-UCB and SMS-EGO are both based on UCB-style hypervolume criteria. Comparing them is therefore important for demonstrating the value of explicitly incorporating the cardinality constraint in MOBO-CAPS.

All algorithms are implemented in \texttt{BoTorch}~\citep{balandat2020botorch}.
Each run begins with $\min(3d,10)$ random initial evaluations.
For all methods, $K$ independent GPs are trained for the $K$ objectives using a Matérn 5/2
kernel with Automatic Relevance Determination (ARD).
Acquisition functions are optimized using multi-start L-BFGS with 12 restarts and 256 raw initial samples, except for HV-KG, for which we use 1 restart following standard \texttt{BoTorch} practice\footnote{\href{https://archive.botorch.org/tutorials/decoupled_mobo}{\text{https://archive.botorch.org/tutorials/decoupled\_mobo}}} due to their high computational cost. The Monte Carlo sample size is fixed at 256.
To choose starting points for L-BFGS, we adopt the Sample-around-Pareto Initialization (SPI). We implemented SPI in \texttt{BoTorch} by setting \texttt{sample\_around\_best=True}, which draws additional candidates in the vicinity of the predicted Pareto set. 
This strategy is inspired by Random Axis-Aligned Subspace Perturbation (RAASP)~\citep{papenmeier2025understanding, rashidi2024cylindrical}, but adapted to the MOBO setting. 
We found that SPI substantially improves REHVI performance; an ablation study is provided in Appendix~\ref{raasp_ablation}.
For fairness, the same initialization is used for CHO-UCB and all baseline methods. We further show that SPI generally does not degrade post-hoc MOBO and CHO-UCB performance, see Appendix~\ref{raasp_ablation} for details. 
We set the CHO-UCB hyperparameter~$\beta_t=4$.

For each algorithm, the incumbent CAPS-set is updated at every iteration. Note that the computational cost of identifying the CAPS-set increases with $M$. We employ exact algorithms to accelerate this step. In the bi-objective case, we use a dynamic programming method that returns the exact optimum in polynomial time~\citep{bringmann2012approximating}. For problems with more than two objectives, we use an exact Branch-and-Bound algorithm~\citep{gomes2018implicit} to find a global optimum.

We measure performance via simple regret, defined as the difference between the true
maximum hypervolume and the best hypervolume obtained by the algorithm so far.
The true maximum hypervolume is approximated
via multi-restart L-BFGS optimization on the $M \times d$-dimensional search space. 
We repeat each experiment 20 times independently.
Results are reported as mean $\pm$ one standard error, and for readability, we apply a
logarithmic scale to the $y$-axis where appropriate; values smaller than $10^{-6}$ are
clipped to $10^{-6}$ to ensure numerical stability. 

Due to space constraints, the main paper reports the \(M=3\) results for all nine test problems. 
Table~\ref{rank_results} summarizes the average ranks, computed from the mean simple regret, across all values of \(M\) and all test problems. 
The complete results are provided in Figure~\ref{full_plot} and Table~\ref{full rank} in Appendix~\ref{apendix_experiment}.

 \begin{figure}[H]
	\centering
  \includegraphics[width=0.86\linewidth]{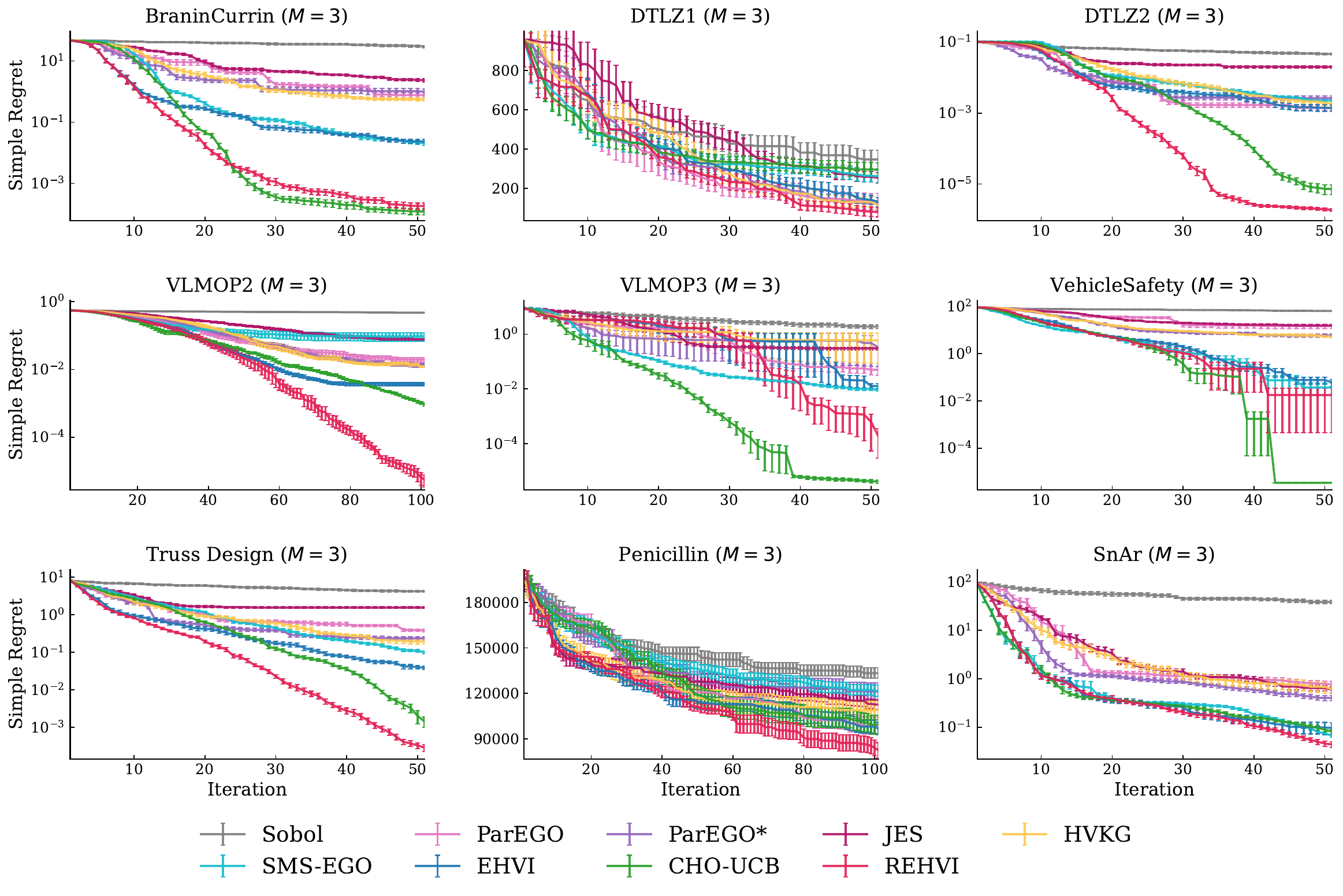}
    \vspace*{-1mm}
	\caption{The cardinality-aware methods, REHVI and CHO-UCB, generally achieve the strongest performance among the compared methods.}
	\label{total_exp_combined_final}
\end{figure}

\begin{table}[t]
\centering
\caption{Average rank of each algorithm across all benchmarks. Lower is better. The first row reports the average over all 45 settings, while the remaining rows report averages separately for each cardinality \(M\). Standard errors are shown in parentheses.}
\label{rank_results}
\resizebox{\linewidth}{!}{%
\begin{tabular}{@{}lccccccccc@{}}
\toprule
 & \multicolumn{2}{c}{\textbf{MOBO-CAPS Methods}} & \multicolumn{7}{c}{\textbf{Baselines}} \\
\cmidrule(lr){2-3} \cmidrule(lr){4-10}
Setting & \textbf{REHVI} & \textbf{CHO-UCB} & EHVI &  SMS-EGO & HV-KG &  ParEGO* & ParEGO & JES & Sobol \\
\midrule
Overall & $\mathbf{1.58}$ $(0.20)$ & $\mathbf{2.51}$ $(0.30)$ & $3.40$ $(0.12)$ & $5.11$ $(0.31)$ & $5.13$ $(0.21)$ & $5.49$ $(0.23)$ & $5.62$ $(0.25)$ & $7.22$ $(0.13)$ & $8.93$ $(0.07)$ \\
\midrule
$M=1$ & $\mathbf{2.00}$ $(0.65)$ & $\mathbf{2.67}$ $(0.85)$ & $3.78$ $(0.32)$ & $5.67$ $(0.71)$ & $4.89$ $(0.51)$ & $4.89$ $(0.63)$ & $5.11$ $(0.65)$ & $7.33$ $(0.37)$ & $8.67$ $(0.33)$ \\
$M=2$ & $\mathbf{1.11}$ $(0.11)$ & $\mathbf{2.67}$ $(0.58)$ & $3.33$ $(0.24)$ & $5.33$ $(0.76)$ & $5.22$ $(0.60)$ & $5.33$ $(0.37)$ & $5.78$ $(0.55)$ & $7.22$ $(0.28)$ & $9.00$ $(0.00)$ \\
$M=3$ & $\mathbf{1.33}$ $(0.17)$ & $\mathbf{2.67}$ $(0.75)$ & $3.56$ $(0.24)$ & $4.89$ $(0.79)$ & $5.11$ $(0.56)$ & $5.89$ $(0.45)$ & $5.56$ $(0.58)$ & $7.00$ $(0.33)$ & $9.00$ $(0.00)$ \\
$M=4$ & $\mathbf{2.11}$ $(0.75)$ & $\mathbf{2.11}$ $(0.54)$ & $3.11$ $(0.20)$ & $4.89$ $(0.75)$ & $5.44$ $(0.18)$ & $5.67$ $(0.62)$ & $5.56$ $(0.58)$ & $7.11$ $(0.26)$ & $9.00$ $(0.00)$ \\
$M=5$ & $\mathbf{1.33}$ $(0.17)$ & $\mathbf{2.44}$ $(0.73)$ & $3.22$ $(0.28)$ & $4.78$ $(0.49)$ & $5.00$ $(0.47)$ & $5.67$ $(0.50)$ & $6.11$ $(0.51)$ & $7.44$ $(0.18)$ & $9.00$ $(0.00)$ \\

\bottomrule
\end{tabular}%
}
\end{table}

Figure~\ref{total_exp_combined_final} shows that REHVI and CHO-UCB achieve the strongest overall performance in the benchmark suite, although the advantage varies. 
This is confirmed by the average ranks in Table~\ref{rank_results}, where REHVI leads, followed by CHO-UCB and then EHVI with post-hoc selection. 


According to in Figure~\ref{full_plot} in Appendix~\ref{Full Experimental Results}, the performance gap between MOBO-CAPS and standard MOBO with post-hoc selection generally becomes smaller as $M$ increases, although this trend is not consistent across all test problems. One possible explanation is that, as $M$ increases, the cardinality constraint becomes less restrictive, potentially reducing the benefit of explicitly optimizing the composition of the solution set.

Overall, REHVI and CHO-UCB outperform standard MOBO with post-hoc selection on average across the benchmark suite. REHVI achieves the strongest overall performance, ranking first in 30 out of 45 settings. This advantage is consistent with the relative strength of the underlying acquisitions: EHVI generally outperforms SMS-EGO, and its replacement-aware counterpart REHVI similarly tends to outperform CHO-UCB. To quantify this relationship, we test whether the pairwise ordering between EHVI and SMS-EGO agrees with that between REHVI and CHO-UCB more often than chance. 
A two-sided binomial test rejects the null hypothesis of random agreement, yielding \(p=2.4589\times10^{-3}\).

We also evaluate DTLZ2 across different input dimensions (3D, 6D, and 10D) and different objective numbers (2,3 and 4 objectives), see Figure~\ref{DTLZ2_M3_Combined} in Appendix~\ref{case_study}. We observe that the relative performance ranking among algorithms remains broadly consistent across all dimensionalities and all objective numbers. In addition, we test DTLZ2 for $M=10,15$ (Figure~\ref{large_M} in Appendix~\ref{case_study}). 
Again we observe that the relative advantage of CHO-UCB and REHVI over baseline acquisition functions decreases as $M$ increases.

 \begin{figure}[H]
	\centering
 \includegraphics[width=0.75\linewidth]{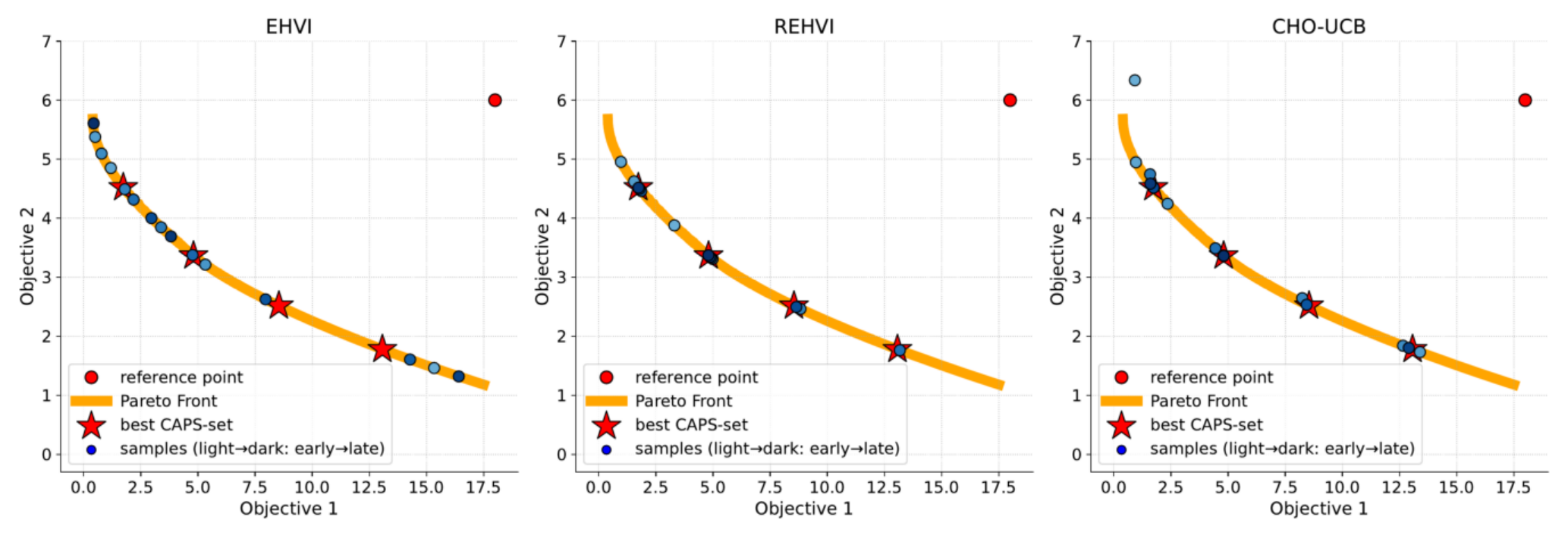}
    \vspace*{-2mm}
	\caption{Examples of sampling behavior of acquisition functions on BraninCurrin ($M=4$)}
	\label{af_vis}
\end{figure}

To better understand the behavior of different acquisition functions, we plot the points sampled by EHVI, CHO-UCB, and REHVI. Figure~\ref{af_vis} visualizes the sampling behavior on BraninCurrin (\(M=4\)). EHVI spreads evaluations along the Pareto front, while CHO-UCB and REHVI concentrate around the best size-\(M\) solutions; REHVI is the most exploitative among the three.

We also compare the wall-clock times of REHVI and CHO-UCB acquisition optimization, see Figure~\ref{wall_time} in Appendix~\ref{appendix_time}. For the values of \(M\) in our experiments, CHO-UCB is faster to optimize than REHVI.

Overall, we recommend REHVI for practical applications. Nevertheless, CHO-UCB remains attractive because its batched formulation admits a formal regret bound, and its acquisition function is computationally cheaper than REHVI.

\paragraph{Methodological Caveats.}
\textcolor{black}{When interpreting these empirical results, two methodological caveats should be noted. First, regarding our statistical significance reporting for the binomial test, the 45 evaluated settings are not strictly independent because they use the same set of test problems for different cardinalities, thus the reported statistical significance may be slightly inflated. Second, the calculation of regret relies on approximating the true maximum hypervolume using multi-start L-BFGS. While this approximation may occasionally yield suboptimal targets, potentially compressing the absolute regret values, this affects all evaluated methods equally. Therefore, while the absolute regret values are subject to this approximation error, the relative ordinal rankings of the algorithms and the resulting comparative analyses remain robust.}

\paragraph{Limitations.}
Our method excels when the surrogate model delivers accurate predictions with well-calibrated uncertainty, as this allows the cardinality-aware acquisition to make reliable selections.
On problems where the Pareto front is difficult to identify within the given evaluation budget (e.g., DTLZ1), the posterior-induced approximation may remain coarse, and explicitly reasoning about CAPS yields limited additional benefit over standard hypervolume-based approaches. 
Consequently, the performance gains of our method are less consistent on such challenging benchmarks.

Our main experiments focus on relatively small cardinalities (\(M\leq5\)), low objective dimensions (\(K\leq3\)), and moderate input dimensions (\(d\leq10\)). Although the DTLZ2 case study provides preliminary evidence for larger \(M\) and \(K\), scalability to larger cardinalities, many-objective settings, and high-dimensional input spaces remains to be studied.

Addressing these limitations represents a valuable avenue for future work.

\section{Conclusion}
We have argued that for many practical applications, the traditional goal of Multi-objective Bayesian Optimization, i.e. approximating the entire Pareto front, is misaligned with the common end-goal of selecting a small, optimal set of solutions. To address this, we introduced a new problem setting, Multi-objective Bayesian Optimization with Cardinality-Aware Pareto Selection (MOBO-CAPS), which directly incorporates the cardinality constraint into the optimization process.

Our proposed algorithms, Replacement-aware Expected Hypervolume Improvement (REHVI) and Collective Higher-dimensional Optimization with Upper Confidence Bound (CHO-UCB), are designed specifically for this setting. Extensive experiments demonstrated that by making the optimization process cardinality-aware, our methods consistently outperform strong baselines that rely on post-hoc selection. We found REHVI to be a powerful empirical performer, while CHO-UCB offers a computationally efficient alternative with favorable theoretical guarantees.

This work opens several promising avenues for future research. Extending REHVI to support batch evaluations and developing a batched version of CHO-UCB with a batch size independent of $M$ would enhance its practical utility. On the methodological front, developing scalable surrogate models that respect the permutation invariance of solution sets could further improve the performance of collective optimization strategies like CHO-EI and CHO-UCB. Finally, moving beyond hypervolume to incorporate other decision-maker preferences, such as fairness or diverse coverage criteria, represents a valuable next step in making Bayesian optimization a more effective tool for real-world decision-making.

\subsubsection*{Acknowledgments}
We sincerely thank the anonymous reviewers for their valuable comments and constructive suggestions, which have greatly helped improve the quality of this paper. The first author gratefully acknowledges support by the Engineering and Physical Sciences Research Council through the Mathematics of Systems II Centre for Doctoral Training at the University of Warwick (reference EP/S022244/1).

\bibliography{ref.bib}
\bibliographystyle{plainnat}
\newpage

\appendix

\section{Appendix}

\subsection{Algorithm Pseudocodes}
\label{Pseudocode}

\begin{algorithm}[H] 
   \caption{REHVI}
   \label{algo:rehvi} 
\begin{algorithmic}
   \STATE Initial observation $\mathcal{D}_0$, initial Pareto set $\mathcal{P}_0$, initial CAPS-set $\mathcal{P}^M_0$
   \FOR{$t = 0, 1, \dots, T-1$}
       \STATE Train $K$ independent GPs $\hat{\mathbf{f}}_t$ given $\mathcal{D}_t$
       \IF{$|\mathcal{P}_t|<M$}
           \STATE $x_{t+1}= \arg\max_{x \in \mathcal{X}}\{ \text{EHVI}(x;\mathcal{P}_t,\hat{\mathbf{f}}_t) \}$
       \ELSE
           \STATE $x_{t+1}= \arg\max_{x \in \mathcal{X}}\{ \text{REHVI}(x;\mathcal{P}^M_t,\hat{\mathbf{f}}_t) \}$
       \ENDIF
       \STATE Query objective function: ${y}_{t+1} = \mathbf{f}(x_{t+1})$
       \STATE Update $\mathcal{D}_{t+1} = \mathcal{D}_{t} \cup \{x_{t+1}, {y}_{t+1}\}$
       \STATE Update the Pareto set $\mathcal{P}_{t+1}$ and the CAPS-set $\mathcal{P}^M_{t+1}$
   \ENDFOR
   \STATE \textbf{return} the CAPS-set $\mathcal{P}^M_{T}$
\end{algorithmic}
\end{algorithm}

\begin{algorithm}[H]
   \caption{CHO-UCB with EHVI Scheduler}
   \label{algo:choucb} 
\begin{algorithmic}[1]
   \STATE Initial observation $\mathcal{D}_0$, initial Pareto set $\mathcal{P}_0$, initial CAPS-set $\mathcal{P}_M^0$
   \FOR{$t = 0, 1, \dots, T-1$}
       \STATE Train $K$ independent GPs $\hat{\mathbf{f}}_t$ given $\mathcal{D}_t$
         \STATE ${X}_{t+1}= \arg\max_{{X} \in \mathcal{X}^{M}}\{ \text{CHO-UCB}({X};\hat{\mathbf{f}}_t) \}~$ (construct an optimistic size-$M$ candidate set)
\STATE $x_{t+1}= \arg\max_{x \in X_{t+1}}\{ \text{EHVI}(x;\mathcal{P}_t,\mathbf{\hat{f}}_t) \} ~$ (select one point using the EHVI scheduler)
       \STATE Query objective function: ${y}_{t+1} = \mathbf{f}(x_{t+1})$
       \STATE Update $\mathcal{D}_{t+1} = \mathcal{D}_{t} \cup \{x_{t+1}, {y}_{t+1}\}$
       \STATE Update the Pareto set $\mathcal{P}_{t+1}$ and the CAPS-set $\mathcal{P}^M_{t+1}$
   \ENDFOR
   \STATE \textbf{return} the CAPS-set $\mathcal{P}^M_{T}$
\end{algorithmic}
\end{algorithm}

\subsection{Comparison of Multi-Objective Optimization Problem Settings}
\label{literature_table}

\begin{table}[H] 
\centering
\label{tab:mobo_comparison_academic}
\begin{tabularx}{\textwidth}{C{3cm} C{2.5cm} C{2.5cm} Y}
\toprule
\textbf{Setting} & \textbf{Metric} & \textbf{Cardinality} & \textbf{Objective Function Property} \\
\midrule
\textbf{Traditional MOBO} & Hypervolume & No limit & Black-box \\
\cmidrule(lr){1-4}
\textbf{MOBO-CAPS} & Hypervolume & $M$ & Black-box \\
\cmidrule(lr){1-4}
\textbf{BOPE \citep{lin2022preference}} & Utility Function & 1 & Black-box \\
\cmidrule(lr){1-4}
\textbf{Knee Finding \citep{heidari2022finding}} & Hypervolume & 1 & Black-box \\
\cmidrule(lr){1-4}
\textbf{Coverage BO \citep{maus2025covering}} & Coverage Score & $M ~ (M < K)$ & Black-box \\
\cmidrule(lr){1-4}
\textbf{HIGA-MO \citep{wang2017hypervolume}} & Hypervolume & $M$ & Accessible Gradient \\
\bottomrule
\end{tabularx}
\end{table}

\subsection{Experiment}
\label{apendix_experiment}
In this section, we provide supplementary material, including detailed descriptions of the benchmark optimization problems and baseline algorithms, implementation details of the proposed methods, computational complexity analysis, and a case study examining the effects of input dimension, number of objectives, and cardinality \(M\).

\subsubsection{Test Problem}
\label{Real-life Test Problems}
In this section, we provide detailed specifications for the benchmark optimization problems used in our evaluation. Note that reference points were selected following literature or standard \texttt{BoTorch} implementations.

\textbf{BraninCurrin}: This problem combines the Branin and Currin single-objective functions into a bi-objective formulation that introduces conflicting landscapes with different scales of nonlinearity. The optimization involves $d=2$ design variables and $K=2$ objectives. Complete problem specifications are available in \citet{emmerich2006single}. The reference point is $[-18.,-6.]$.

\textbf{DTLZ1}: This benchmark is a standard multi-objective test function designed to evaluate an algorithm’s ability to converge toward and maintain a uniform spread along a linear Pareto front. The optimization involves $d=3$ design variables and $K=2$ objectives. Complete problem specifications are available in \citet{deb2005scalable}. The reference point is $[-50., -50.]$.

\textbf{DTLZ2}: This test function features a nonlinear, convex Pareto front and is widely used to assess convergence accuracy and diversity maintenance in multi-objective optimization algorithms. The optimization involves $d=3$ design variables and $K=2$ objectives. Complete problem specifications are available in \citet{deb2005scalable}. The reference point is $[-1.1,-1.1]$.

\textbf{VLMOP2}: This benchmark consists of two smooth, nonlinear objective functions that induce a convex Pareto front, making it suitable for evaluating convergence behavior in bi-objective optimization. The optimization involves $d=6$ design variables and $K=2$ objectives. Complete problem specifications are available in \citet{van1999multiobjective}. The reference point is $[-1.2, -1.2]$.

\textbf{VLMOP3}: This benchmark presents two smooth, nonlinear objectives that generate a concave Pareto front, providing a moderate challenge for both convergence and diversity. The optimization involves $d=2$ design variables and $K=3$ objectives. Complete problem specifications are available in \citet{van1999multiobjective}. The reference point is $[-4.,-25.,-0.25]$.

\textbf{Truss Design}: This optimization problem involves designing a four-bar truss structure with 2 competing objectives: minimizing structural volume while minimizing joint displacement. The design variables correspond to the lengths of the four individual bars. Complete problem specifications are available in \citet{cheng1999generalized}. The reference point is $[-2886.37, -0.04]$.

\textbf{SnAr}: This optimization problem involves a nucleophilic aromatic substitution reaction (SnAr) between 2,4-difluoronitrobenzene and pyrrolidine in ethanol. The optimization uses 4 design variables representing initial conditions to optimize space-time yield and environmental impact.  Complete problem specifications are available in \citet{hone2017rapid}. The reference point is $[0., 120.]$.

\textbf{VehicleSafety}: This multi-objective optimization problem involves designing vehicle safety systems with competing objectives of optimizing mass, collision acceleration, and toe board intrusion by designing the thickness of five reinforced members around the frontal structure. The optimization uses 5 design variables representing the thickness parameters of structural components to simultaneously optimize 3 conflicting objectives. Complete problem specifications are available in \citep{liao2008multiobjective}. The reference point is $[1864.7,-11.82,-0.29]$.

\textbf{Penicillin}: This benchmark optimizes the Penicillin production process. The optimization involves 7 design variables controlling initial reaction conditions to optimize 3 objectives: yield, carbon dioxide emissions, and fermentation time.  Complete problem specifications are available in \citet{liang2021scalable}. The reference point is $[ -1.85,  -86.93, -514.7]$.

\subsubsection{CAPS-set Coverage of the Pareto Front}

\begin{figure}[H]
	\centering
  \includegraphics[width=0.95\linewidth]{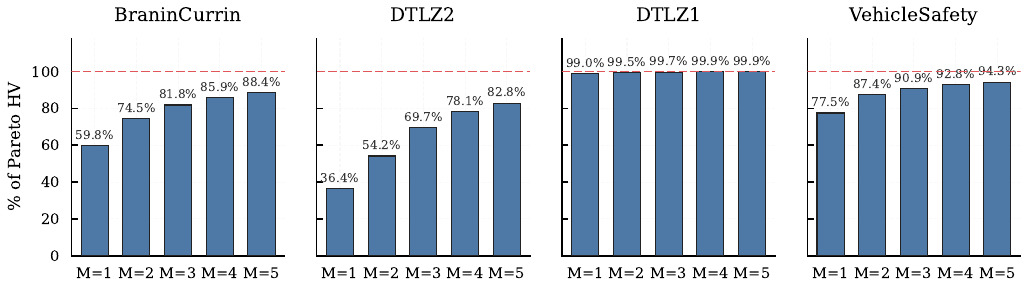}
    \vspace*{-1mm}
	\caption{Percentage of Pareto Hypervolume Covered by the CAPS-set}
	\label{hv_percentage_bar}
\end{figure}

We also report the percentage of Pareto-front hypervolume covered by the CAPS-set in Figure~\ref{hv_percentage_bar}. Even for \(M=1\), the CAPS-set covers at least \(36.4\%\) of the Pareto-front hypervolume, while for \(M=5\), it covers more than \(80\%\) in all cases.

\subsubsection{Benchmark Algorithm}
\label{Benchmark Algorithm}
We compare the proposed methods against a diverse set of benchmark algorithms that are applicable to the considered problems. Below, we briefly summarize each method. \\\\
\textbf{Sobol:} Sobol is a quasi-random search strategy based on low-discrepancy sequences. It does not use a surrogate model and serves as a simple model-free baseline. At each iteration, a new design point is selected from a Sobol sequence within the search domain.\\\\
\textbf{EHVI:}
EHVI \citep{emmerich2008computation,daulton2020differentiable} is a standard acquisition function for multi-objective Bayesian optimization. Given the current set of observed non-dominated points, EHVI selects the candidate that maximizes the expected increase in hypervolume after one additional evaluation. EHVI is designed to improve the approximation of the entire Pareto front rather than to directly optimize a fixed-cardinality subset.\\\\
\textbf{SMS-EGO:}
SMS-EGO~\citep{ponweiser2008multiobjective} is a sequential model-based multi-objective optimization method that extends the single-objective EGO framework to the multi-objective setting. It uses upper confidence bound to calculate the hypervolume, in order to identify the next point for evaluation. Similar to EHVI, its goal is to improve the global Pareto-front approximation rather than to optimize a set of exactly \(M\) solutions. In our experiments, we set $\beta_t=4$ in SMS-EGO.\\\\
\textbf{ParEGO:}
ParEGO \citep{knowles2006parego} scalarizes the multiple objectives into a single objective using a randomly sampled weight vector and then applies standard single-objective Bayesian optimization to the scalarized problem. By varying the scalarization over iterations, ParEGO is able to explore different parts of the Pareto front. However, it does not explicitly account for the cardinality constraint in the final recommended solution set.\\\\
\textbf{\(\text{ParEGO}^*\):}
\(\text{ParEGO}^*\) denotes a reference-point variant of ParEGO that incorporates the hypervolume reference point into the scalarization or recommendation procedure. This variant is included to provide a stronger scalarization-based baseline that is better aligned with hypervolume-oriented evaluation. Nevertheless, like ParEGO, it does not directly optimize a fixed-size subset of Pareto solutions.\\\\
\textbf{JES:}
Joint Entropy Search (JES)~\citep{tu2022joint} is an information-theoretic acquisition function that selects evaluations to maximally reduce uncertainty about the Pareto-optimal set and Pareto front. In the multi-objective setting, JES seeks to efficiently learn the Pareto front by maximizing information gain. In our experiments, we set the number of Pareto set samples to $M$ and the number of Pareto points per sample to $M$. All remaining implementation details follow the standard \texttt{BoTorch} tutorial.\\\\
\textbf{HV-KG:}
Hypervolume Knowledge Gradient (HV-KG)~\citep{daulton2023hypervolume} is a one-step lookahead acquisition function that evaluates the expected value of an observation through its effect on the posterior-mean hypervolume after conditioning on fantasy outcomes. HV-KG is especially relevant to MOBO-CAPS because its terminal recommendation set has a fixed size, denoted \(N_p\) in the original formulation. We set \(N_p=M\), so that the terminal decision in HV-KG is the best posterior-mean size-\(M\) set. This makes HV-KG a strong lookahead baseline for the CAPS objective. Unless otherwise stated, the remaining implementation details follow the standard \texttt{BoTorch} implementation.

\subsubsection{Full Experimental Results}
\label{Full Experimental Results}
\begin{figure}[htbp]
	\centering
 \includegraphics[width=0.9\linewidth]{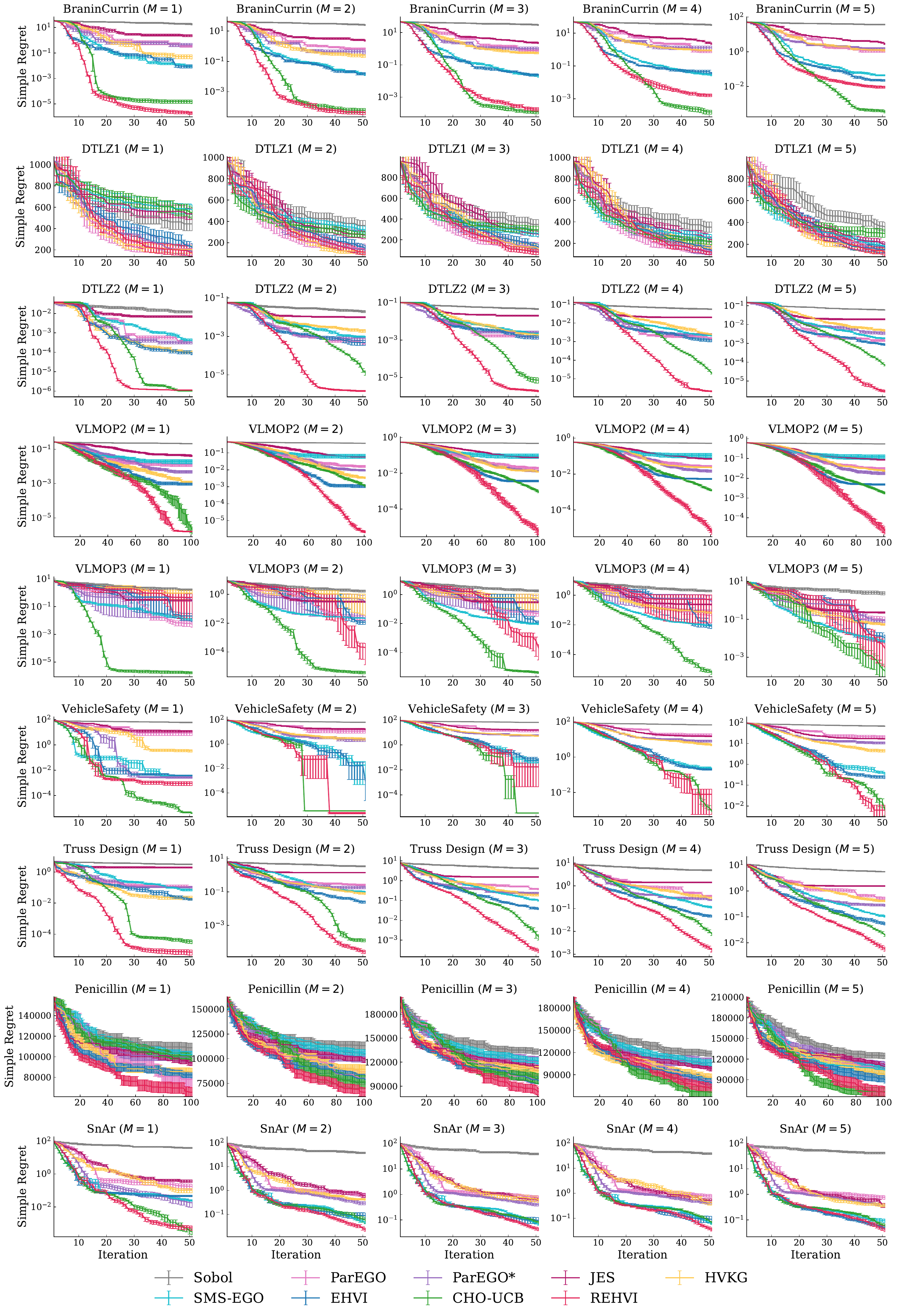}
    \vspace*{-1mm}
	\caption{The cardinality-aware methods, REHVI and CHO-UCB, achieve the strongest overall performance across settings.}
        \label{full_plot}
\end{figure}

Figure~\ref{full_plot} reports the complete results across all test problems and all values of $M$. Table~\ref{full rank} presents the ranking of each algorithm based on its final regret mean for each test problem. Overall, the strongest performance is achieved by the MOBO-CAPS methods, namely REHVI and CHO-UCB.

\label{full_experimenta_result}

\begin{table}[H]
\centering
\caption{Ranked performances (lower is better) of all algorithms for all benchmarks}
\label{full rank}
\resizebox{\textwidth}{!}{%
\begin{tabular}{lccccccccc}
\toprule
 & {Sobol} & {ParEGO} & {ParEGO*} & {JES} & {HV-KG} & {SMS-EGO} & {EHVI} & {CHO-UCB} & {REHVI} \\
\midrule
BraninCurrin ($M=1$) & $9$ & $6$ & $7$ & $8$ & $5$ & $4$ & $3$ & $2$ & $\mathbf{1}$ \\
BraninCurrin ($M=2$) & $9$ & $7$ & $6$ & $8$ & $5$ & $3$ & $4$ & $2$ & $\mathbf{1}$ \\
BraninCurrin ($M=3$) & $9$ & $6$ & $7$ & $8$ & $5$ & $3$ & $4$ & $\mathbf{1}$ & $2$ \\
BraninCurrin ($M=4$) & $9$ & $5$ & $7$ & $8$ & $6$ & $3$ & $4$ & $\mathbf{1}$ & $2$ \\
BraninCurrin ($M=5$) & $9$ & $5$ & $7$ & $8$ & $6$ & $4$ & $3$ & $\mathbf{1}$ & $2$ \\
\midrule
DTLZ1 ($M=1$) & $6$ & $4$ & $2$ & $7$ & $3$ & $9$ & $5$ & $8$ & $\mathbf{1}$ \\
DTLZ1 ($M=2$) & $9$ & $5$ & $3$ & $6$ & $2$ & $8$ & $4$ & $7$ & $\mathbf{1}$ \\
DTLZ1 ($M=3$) & $9$ & $5$ & $3$ & $6$ & $2$ & $7$ & $4$ & $8$ & $\mathbf{1}$ \\
DTLZ1 ($M=4$) & $9$ & $4$ & $2$ & $7$ & $5$ & $8$ & $3$ & $6$ & $\mathbf{1}$ \\
DTLZ1 ($M=5$) & $9$ & $4$ & $3$ & $7$ & $2$ & $6$ & $5$ & $8$ & $\mathbf{1}$ \\
\midrule
DTLZ2 ($M=1$) & $9$ & $6$ & $5$ & $8$ & $4$ & $7$ & $3$ & $\mathbf{1}$ & $2$ \\
DTLZ2 ($M=2$) & $9$ & $3$ & $5$ & $8$ & $7$ & $6$ & $4$ & $2$ & $\mathbf{1}$ \\
DTLZ2 ($M=3$) & $9$ & $3$ & $7$ & $8$ & $5$ & $6$ & $4$ & $2$ & $\mathbf{1}$ \\
DTLZ2 ($M=4$) & $9$ & $4$ & $7$ & $8$ & $6$ & $5$ & $3$ & $2$ & $\mathbf{1}$ \\
DTLZ2 ($M=5$) & $9$ & $4$ & $6$ & $8$ & $7$ & $5$ & $3$ & $2$ & $\mathbf{1}$ \\
\midrule
VLMOP2 ($M=1$) & $9$ & $6$ & $5$ & $8$ & $4$ & $7$ & $3$ & $2$ & $\mathbf{1}$ \\
VLMOP2 ($M=2$) & $9$ & $6$ & $5$ & $7$ & $4$ & $8$ & $2$ & $3$ & $\mathbf{1}$ \\
VLMOP2 ($M=3$) & $9$ & $6$ & $5$ & $7$ & $4$ & $8$ & $3$ & $2$ & $\mathbf{1}$ \\
VLMOP2 ($M=4$) & $9$ & $6$ & $4$ & $7$ & $5$ & $8$ & $3$ & $2$ & $\mathbf{1}$ \\
VLMOP2 ($M=5$) & $9$ & $6$ & $4$ & $7$ & $5$ & $8$ & $3$ & $2$ & $\mathbf{1}$ \\
\midrule
VLMOP3 ($M=1$) & $9$ & $2$ & $5$ & $6$ & $8$ & $3$ & $4$ & $\mathbf{1}$ & $7$ \\
VLMOP3 ($M=2$) & $9$ & $5$ & $6$ & $7$ & $8$ & $4$ & $3$ & $\mathbf{1}$ & $2$ \\
VLMOP3 ($M=3$) & $9$ & $5$ & $7$ & $6$ & $8$ & $3$ & $4$ & $\mathbf{1}$ & $2$ \\
VLMOP3 ($M=4$) & $9$ & $4$ & $7$ & $6$ & $5$ & $3$ & $2$ & $\mathbf{1}$ & $8$ \\
VLMOP3 ($M=5$) & $9$ & $7$ & $6$ & $8$ & $5$ & $3$ & $4$ & $\mathbf{1}$ & $2$ \\
\midrule
VehicleSafety ($M=1$) & $9$ & $7$ & $3$ & $8$ & $6$ & $4$ & $5$ & $\mathbf{1}$ & $2$ \\
VehicleSafety ($M=2$) & $9$ & $7$ & $5$ & $8$ & $6$ & $4$ & $3$ & $2$ & $\mathbf{1}$ \\
VehicleSafety ($M=3$) & $9$ & $7$ & $6$ & $8$ & $5$ & $3$ & $4$ & $\mathbf{1}$ & $2$ \\
VehicleSafety ($M=4$) & $9$ & $8$ & $6$ & $7$ & $5$ & $4$ & $3$ & $\mathbf{1}$ & $2$ \\
VehicleSafety ($M=5$) & $9$ & $8$ & $6$ & $7$ & $5$ & $4$ & $3$ & $\mathbf{1}$ & $2$ \\
\midrule
TrussDesign ($M=1$) & $9$ & $6$ & $7$ & $8$ & $4$ & $5$ & $3$ & $2$ & $\mathbf{1}$ \\
TrussDesign ($M=2$) & $9$ & $7$ & $6$ & $8$ & $4$ & $5$ & $3$ & $2$ & $\mathbf{1}$ \\
TrussDesign ($M=3$) & $9$ & $7$ & $6$ & $8$ & $5$ & $4$ & $3$ & $2$ & $\mathbf{1}$ \\
TrussDesign ($M=4$) & $9$ & $7$ & $5$ & $8$ & $6$ & $4$ & $3$ & $2$ & $\mathbf{1}$ \\
TrussDesign ($M=5$) & $9$ & $7$ & $5$ & $8$ & $6$ & $4$ & $3$ & $2$ & $\mathbf{1}$ \\
\midrule
Penicillin ($M=1$) & $9$ & $2$ & $7$ & $5$ & $4$ & $8$ & $3$ & $6$ & $\mathbf{1}$ \\
Penicillin ($M=2$) & $9$ & $4$ & $7$ & $6$ & $5$ & $8$ & $3$ & $2$ & $\mathbf{1}$ \\
Penicillin ($M=3$) & $9$ & $3$ & $7$ & $6$ & $5$ & $8$ & $2$ & $4$ & $\mathbf{1}$ \\
Penicillin ($M=4$) & $9$ & $4$ & $8$ & $6$ & $5$ & $7$ & $3$ & $\mathbf{1}$ & $2$ \\
Penicillin ($M=5$) & $9$ & $6$ & $8$ & $7$ & $4$ & $5$ & $3$ & $2$ & $\mathbf{1}$ \\
\midrule
SnAr ($M=1$) & $9$ & $7$ & $3$ & $8$ & $6$ & $4$ & $5$ & $\mathbf{1}$ & $2$ \\
SnAr ($M=2$) & $9$ & $8$ & $5$ & $7$ & $6$ & $2$ & $4$ & $3$ & $\mathbf{1}$ \\
SnAr ($M=3$) & $9$ & $8$ & $5$ & $6$ & $7$ & $2$ & $4$ & $3$ & $\mathbf{1}$ \\
SnAr ($M=4$) & $9$ & $8$ & $5$ & $7$ & $6$ & $2$ & $4$ & $3$ & $\mathbf{1}$ \\
SnAr ($M=5$) & $9$ & $8$ & $6$ & $7$ & $5$ & $4$ & $2$ & $3$ & $\mathbf{1}$ \\
\bottomrule
\end{tabular}%
}
\end{table}

\subsubsection{Case Study: Effects of Dimension, Objective Count, and Cardinality}
\label{case_study}
\begin{figure}[H]
	\centering
 \includegraphics[width=0.85\linewidth]{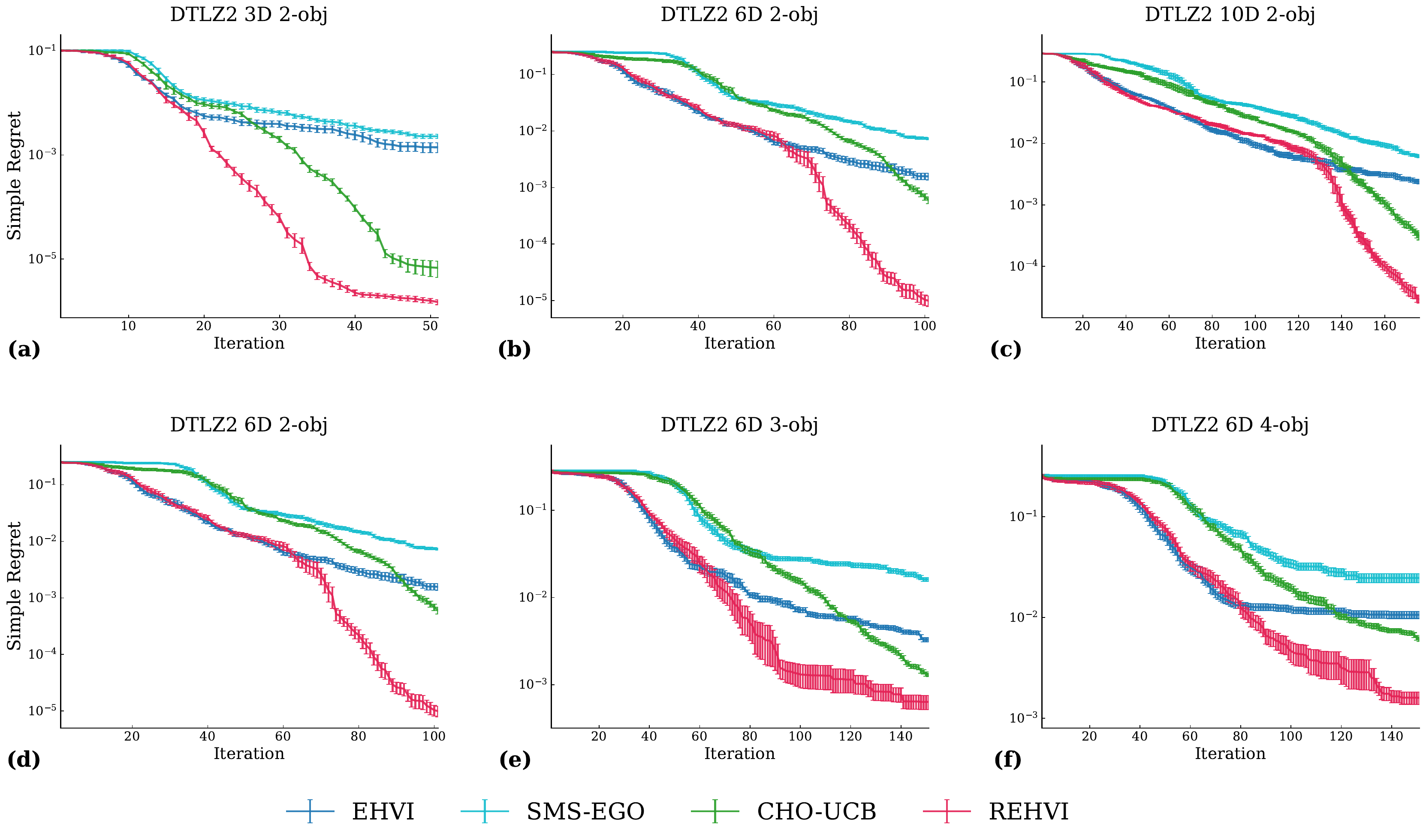}
	\caption{Results on DTLZ2 across different input dimensionalities and different numbers of objectives. The relative ranking among algorithms remains broadly consistent across settings.}
	\label{DTLZ2_M3_Combined}
\end{figure}

In this section, we conduct a case study on DTLZ2 to examine how the performance of MOBO-CAPS algorithms varies with input dimension, number of objectives, and target cardinality. In Figure~\ref{DTLZ2_M3_Combined}, we evaluate DTLZ2 with input dimensions \(d=3,6,10\) for the bi-objective case, and with \(K=2,3,4\) objectives for a fixed input dimension \(d=6\). The relative ranking among algorithms remains broadly consistent across these settings.

We also evaluate two larger cardinalities, \(M=10\) and \(M=15\). The results show that the relative advantage of REHVI and CHO-UCB over baseline acquisition functions decreases as \(M\) increases.

\begin{figure}[H]
	\centering
 \includegraphics[width=0.99\linewidth]{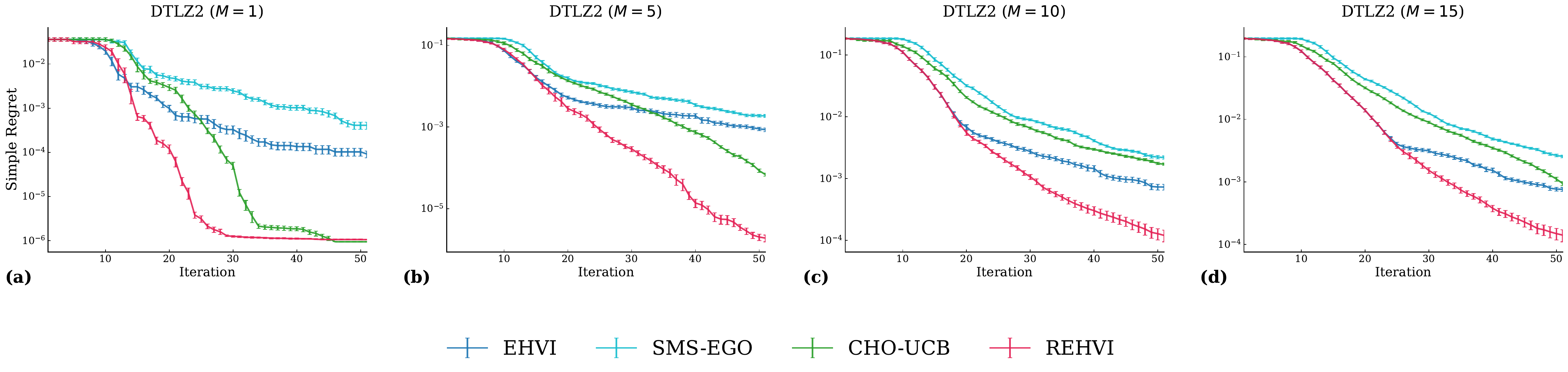}
	\caption{Results on DTLZ2 for larger target cardinalities $M$. The relative advantage of REHVI and CHO-UCB over baseline acquisition functions decreases as $M$ increases.}
	\label{large_M}
\end{figure}

\subsection{Implementation and Complexity Analysis}
\label{appendix_time}

\begin{figure}[H]
	\centering
 \includegraphics[width=0.66\linewidth]{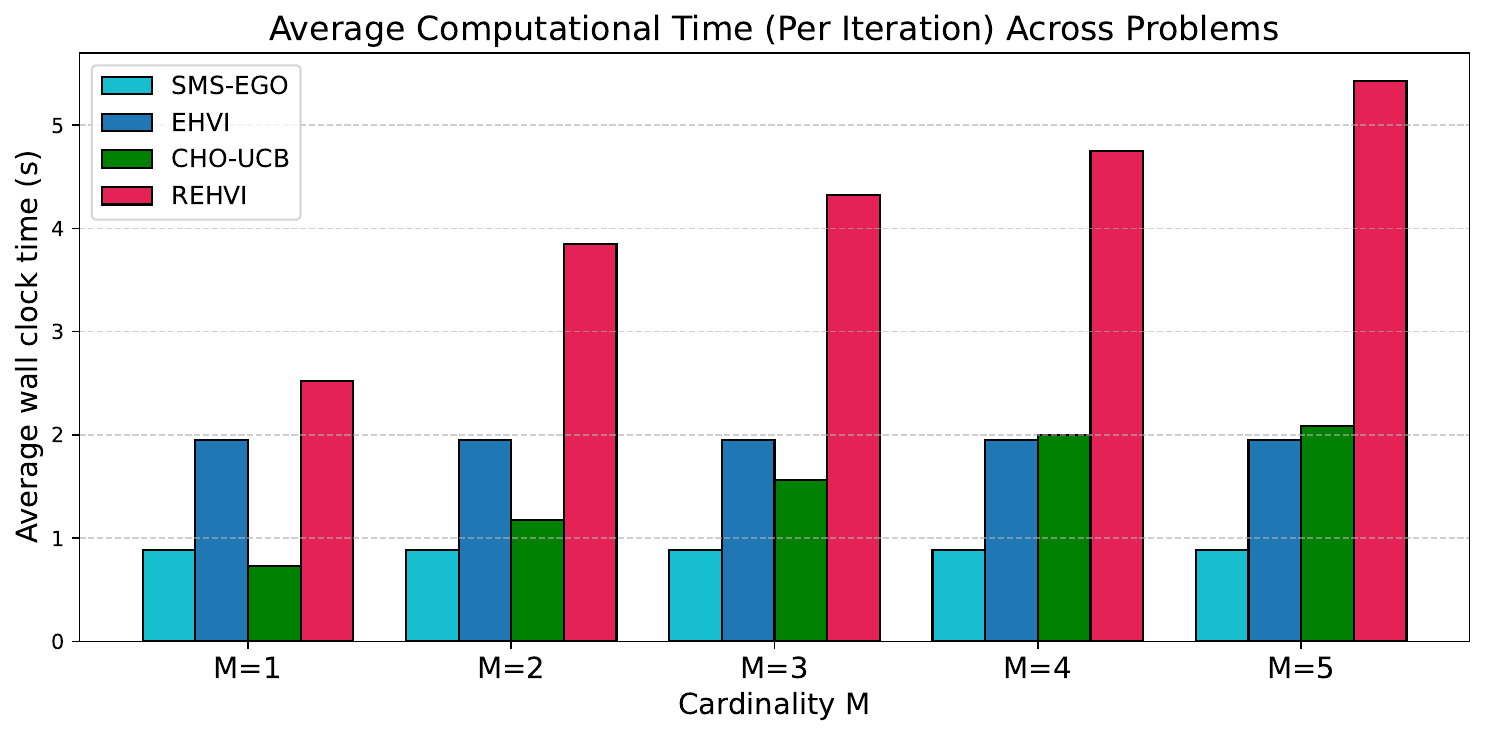}

	\caption{Wall clock time grows with~$M$. CHO-UCB is computationally cheaper than REHVI.}
	\label{wall_time}
\end{figure}

\textcolor{black}{We implement REHVI in \texttt{BoTorch}~\citep{balandat2020botorch}. Specifically, for each candidate \(x\), we draw Monte Carlo samples from the GP posterior and evaluate the corresponding sampled objective values. For each sample, we compute the maximum hypervolume gain achievable by a one-point swap with the current CAPS-set. In the implementation, we add the new candidate point to the current size-\(M\) set and then remove the point whose deletion maximizes the hypervolume of the remaining \(M\) points. We use the best such replacement to compute the replacement-aware hypervolume gain. The REHVI value of \(x\) is then estimated by averaging this maximal gain over posterior samples. This implementation is fully differentiable and supports automatic differentiation. }\\\\
We implement CHO-UCB in \texttt{BoTorch}~\citep{balandat2020botorch}. For each candidate \(x\), we compute the GP posterior mean and standard deviation for each objective, and form an optimistic estimate \(\mu(x) + \sqrt{\beta_t} \sigma(x)\), where \(\beta_t\) controls the exploration-exploitation trade-off. We then compute the hypervolume using the corresponding non-dominated partitioning and reference point. In this way, CHO-UCB favors candidates that are predicted to yield large hypervolume under an upper confidence bound criterion. This implementation is fully differentiable and amenable to gradient-based optimization.\\\\
The time complexity of our proposed methods can be analyzed by starting with the q-Expected Hypervolume Improvement (qEHVI) algorithm, which has a complexity of $O(KNP(2^q-1))$ \citep{daulton2020differentiable}. Here, $K$ is the number of objective functions, $N$ is the number of Monte Carlo samples, $P$ is the number of disjoint hyper-rectangles that result from the decomposition of the non-dominated objective space, and $q$ is the batch size. Furthermore, it is important to note that $P$ is a function of $K$ and the Pareto front size $|\mathcal{P}_t|$. In the two-objective case, $P = |\mathcal{P}_t| + 1$, but $P$ is super-polynomial in $|\mathcal{P}_t|$ for higher $K$ values. We therefore present the complexity analysis for the bi-objective case, where the expression takes a clear form:
\begin{itemize}
    \item \textbf{REHVI:} Our method builds upon EHVI (\(q=1\)) with a fixed-size candidate set \((|\mathcal{P}|=M)\). 
    The acquisition function must iterate through each of the \(M\) points in this set to identify an optimal replacement, 
    which introduces an additional factor of \(M\). 
    This results in a time complexity of \(O(KNM^2)\).

    \item \textbf{CHO-UCB}: This method has a batch size $q=M$ and current Pareto front size $|\mathcal{P}|=0$ but offers a significant computational advantage by eliminating the need for Monte Carlo simulation ($N=1$). Its time complexity is therefore $O(K(2^M-1))$. Although the computational cost grows exponentially with $M$, it remains strictly lower than that of qEHVI ($q=M$), ensuring the method scales effectively for practical values of $M$.
\end{itemize}
\noindent
Figure~\ref{wall_time} shows the average wall clock time of acquisition function optimization of SMS-EGO, EHVI, CHO-UCB and REHVI (surrogate model training time not included). Note that for SMS-EGO and EHVI, we calculate $M=1,2,3,4,5$ at the same time, so the time for different $M$ is the same. All computations were performed on an Intel Core i7-1355U
CPU. For CHO-UCB, and REHVI, we observe that acquisition function optimization time generally increases with $M$. Among the cardinality-aware acquisitions considered, REHVI exhibits the longest wall clock time, while CHO-UCB requires the least computational time as it does not require Monte Carlo simulation. 

As a side note, acquisition function optimization in this setting is relatively inexpensive. The primary computational burden lies in updating the CAPS-set, specifically, identifying the subset of $M$ solutions from $N$ observations that maximizes hypervolume, given that this subset selection is an NP-hard problem.

\subsection{Ablation Study on SPI}
\label{raasp_ablation}
We observe that acquisition function optimization, particularly the choice of
initialization strategy, has a significant effect on the performance of both
REHVI and CHO-EI.
In this section, we perform a detailed analysis of the impact of
SPI on
REHVI, CHO-UCB, and CHO-EI.
Notably, the optimization of the acquisition function proves to be more
challenging for CHO-EI than for the other two methods,
regardless of whether SPI is applied, which likely contributes to its
inferior overall performance.

 \begin{figure}[htbp]
	\centering
 \includegraphics[width=0.81\linewidth]{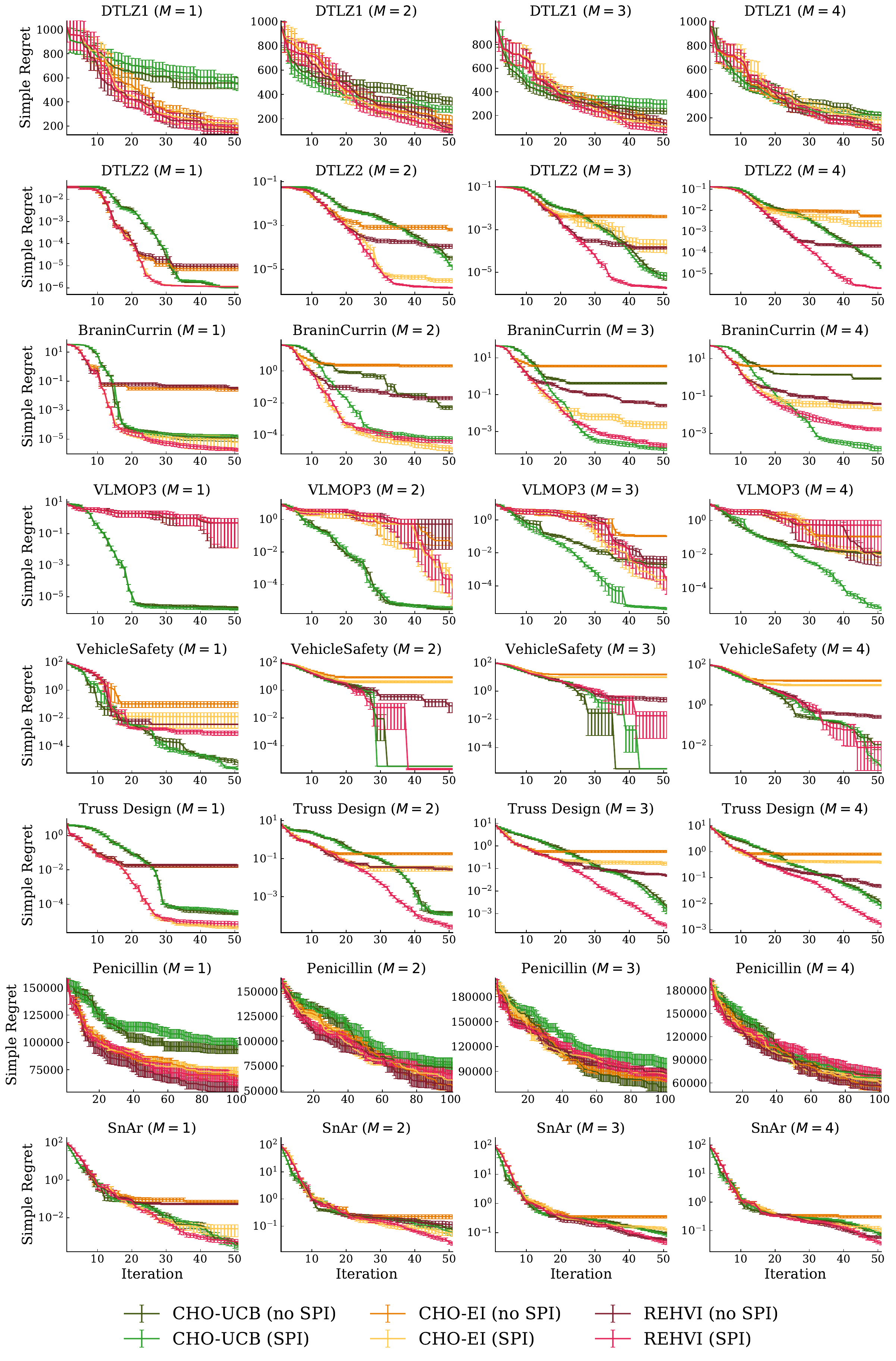}
    \vspace*{-1mm}
	\caption{Ablation study of SPI}
	\label{Ablation_RAASP}
\end{figure}

Figure~\ref{Ablation_RAASP} presents an ablation study illustrating that
SPI substantially improves the performance of REHVI and
CHO-EI, whereas its effect on CHO-UCB is comparatively modest.

\begin{figure}[H]
	\centering
 \includegraphics[width=0.75\linewidth]{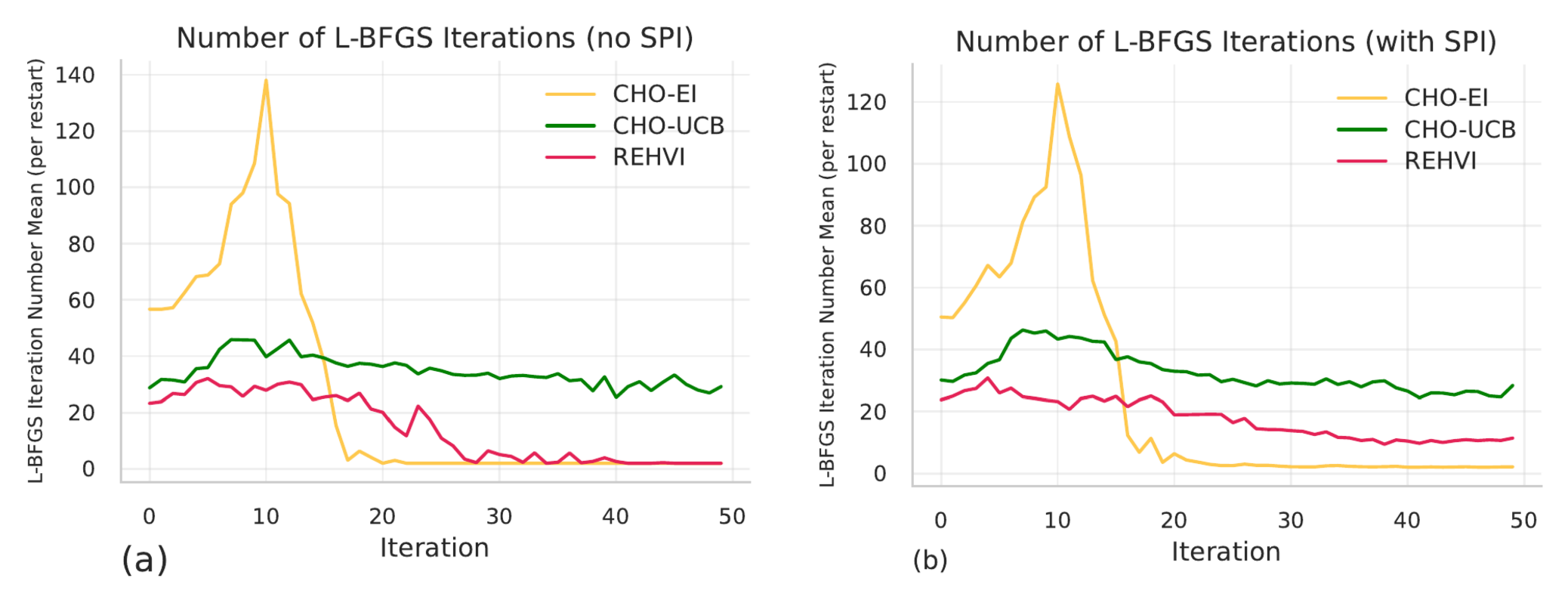}
    \vspace*{-1mm}
	\caption{Number of L-BFGS Iterations}
	\label{acquisition_iter}
\end{figure}

To further investigate how SPI influences acquisition function
optimization, we track the evolution of acquisition values across iterations
with and without SPI on the DTLZ2 test function with $M=3$.
We also analyze the number of L-BFGS iterations per restart as a proxy for
optimization stability: a small iteration count indicates premature
termination, whereas a larger one reflects a smoother and more stable
optimization process.
Figure~\ref{acquisition_iter} compares the two scenarios, i.e. without SPI and
with SPI, demonstrating that SPI yields more stable and reliable
acquisition optimization dynamics.

\begin{figure}[H]
	\centering
 \includegraphics[width=0.75\linewidth]{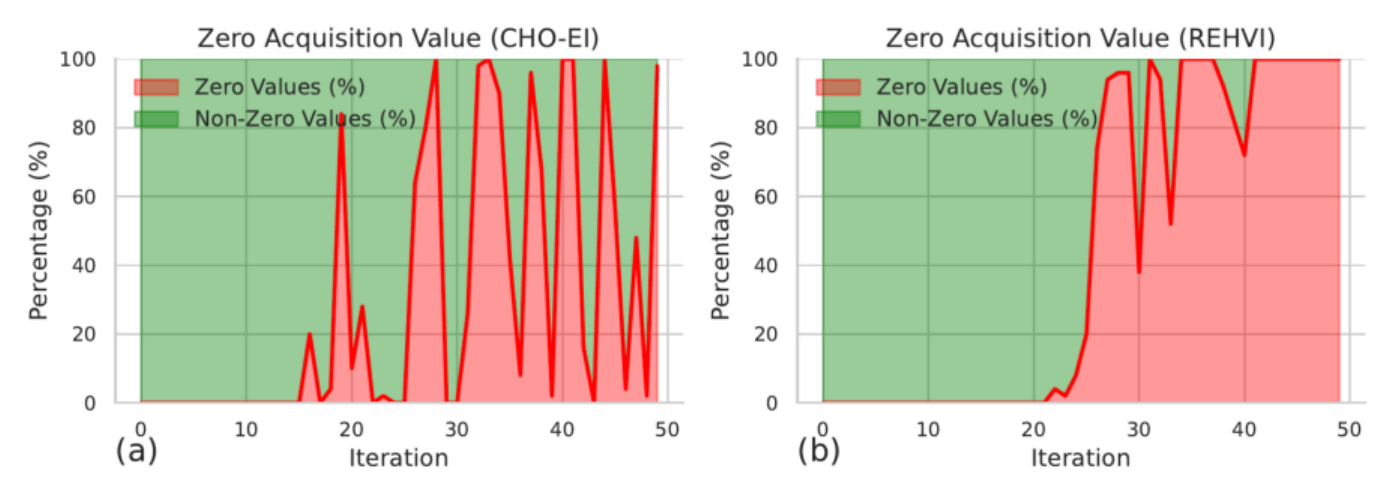}
    \vspace*{-1mm}
	\caption{Percentage of Zero Acquisition Values (no SPI): Red indicates the proportion of zero values; green represents non-zero values. CHO-EI and REHVI exhibit a high percentage of zero values, indicating much invalid acquisition optimization.}
	\label{af_zero_no_raasp}
\end{figure}

\begin{figure}[H]
	\centering
 \includegraphics[width=0.75\linewidth]{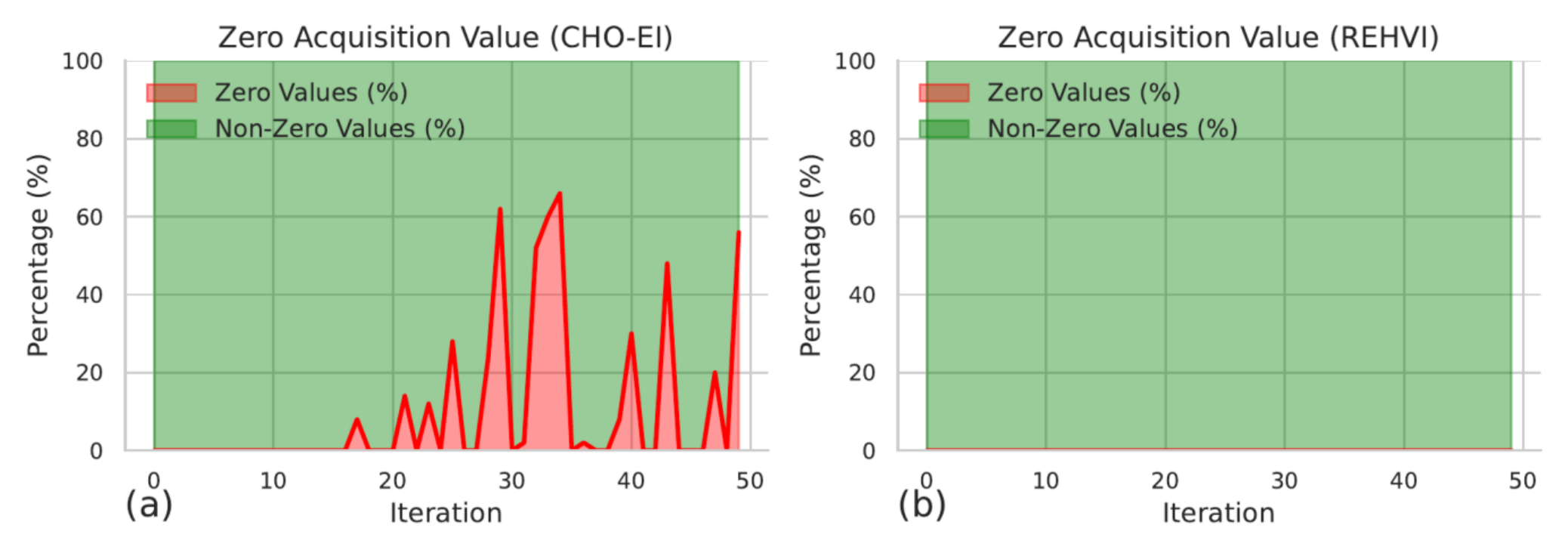}
    \vspace*{-1mm}
	\caption{Percentage of Zero Acquisition Values (with SPI): Red indicates the proportion of zero values; green represents non-zero values.}
	\label{af_zero}
\end{figure}

In Figure~\ref{acquisition_iter} (a), we observe that, for CHO-EI and REHVI, the iteration count rapidly drops to near zero in the later stages. This decline occurs because the initial acquisition function value becomes zero, causing the optimization to terminate almost immediately  returning zero acquisition function value (see Figure~\ref{af_zero_no_raasp}). In Figure~\ref{acquisition_iter} (b), with SPI applied, the iteration numbers become more stable for both REHVI and CHO-EI. However, CHO-EI still experiences a drop in iterations, approaching zero in some cases. This behavior aligns with the percentage of zero acquisition values shown in Figure~\ref{af_zero}. While REHVI no longer encounters zero initial values with SPI, CHO-EI still exhibits some, albeit at a reduced frequency.

\begin{figure}[H]
	\centering
 \includegraphics[width=0.95\linewidth]{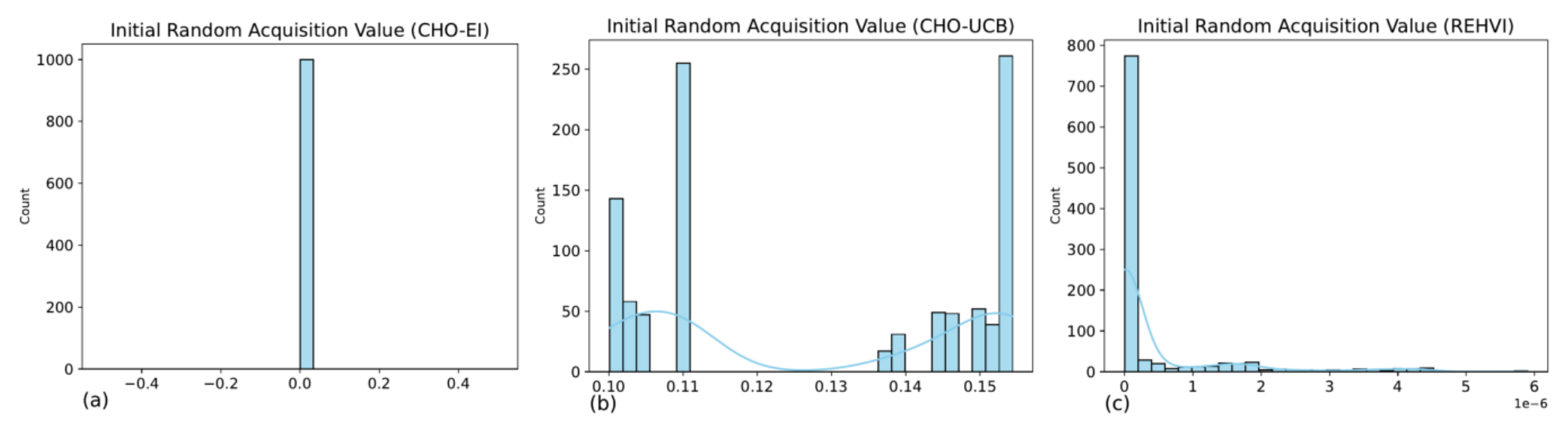}
    \vspace*{-1mm}
	\caption{Initial Random Acquisition Value (with SPI)}
	\label{initial_af}
\end{figure}

We find that for both initialization strategies, CHO-EI performs worse than CHO-UCB and REHVI primarily due to challenges in acquisition function optimization. We observe that CHO-EI exhibits a flatter landscape compared to the other two methods. Consequently, it is more difficult to obtain valid initial acquisition function values to start L-BFGS optimization. In Figure~\ref{initial_af} (a), we sample 1000 points using CHO-EI at the 40th iteration on SPI. However, all sampled points yield zero acquisition function values. This issue is less severe for CHO-UCB and REHVI. As shown in Figures~\ref{initial_af}(b) and (c), the 1000 samples exhibit diverse acquisition function values, making the optimization landscape more amenable to gradient-based methods.

\begin{figure}[H]
	\centering
 \includegraphics[width=0.99\linewidth]{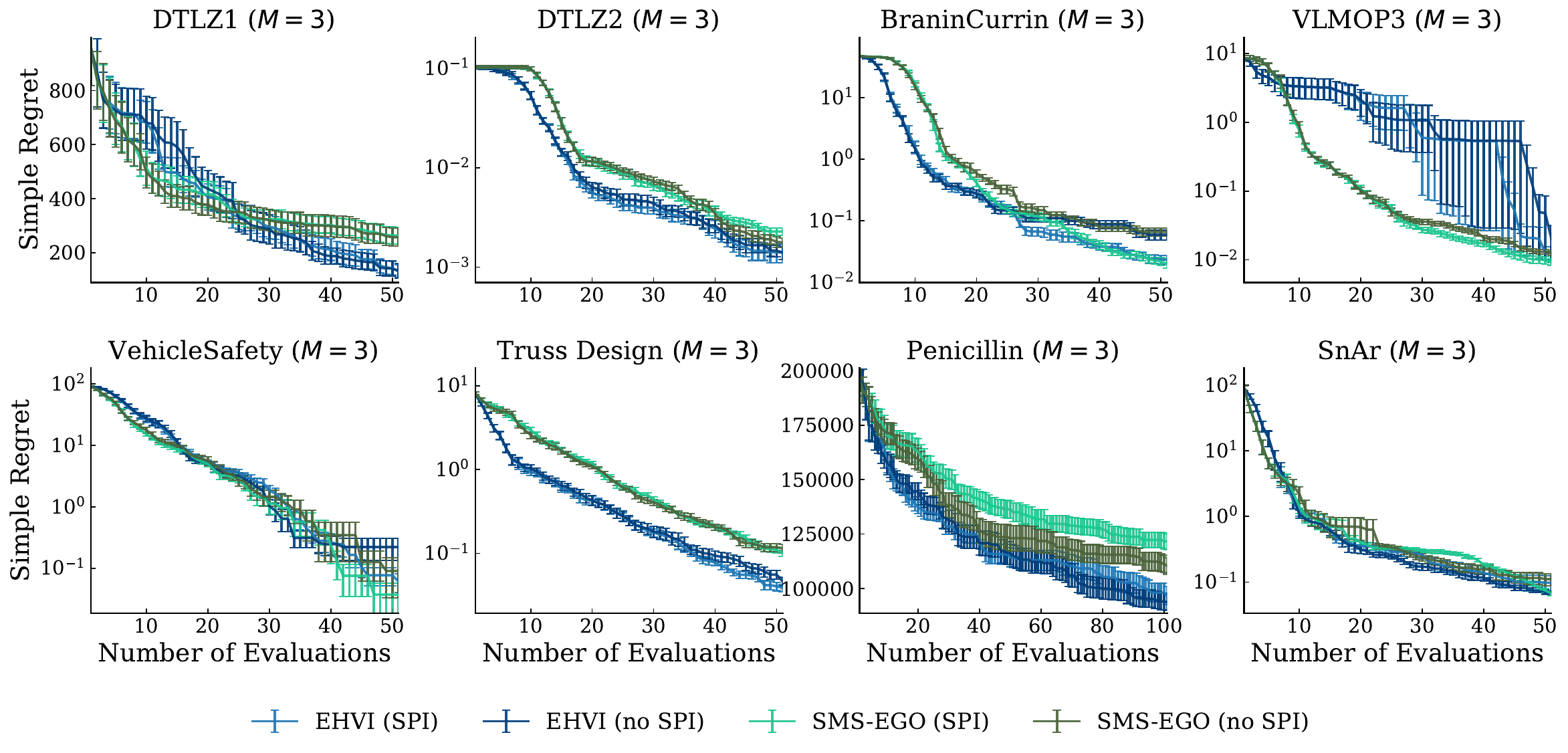}
    \vspace*{-1mm}
	\caption{SPI generally does not degrade post-hoc MOBO performance.}
	\label{normal_ablation}
\end{figure}

In addition, we conduct an ablation study of SPI for MOBO with post-hoc selection (see Figure~\ref{normal_ablation}). Overall, SPI does not degrade performance. The only exception is the Penicillin problem, where SPI reduces performance; however, a similar degradation is also observed for REHVI and CHO-UCB (see Figure~\ref{Ablation_RAASP}).

\subsection{Effect of the Scheduler in CHO-UCB}
\label{scheduler}
We investigate the role of the evaluation scheduler in CHO-UCB for the $M=3$ setting.
\textcolor{black}{We compare five variants of CHO‑UCB: (i) the batched variant \textsc{CHO‑UCB (Batch)}, which evaluates all 
$M$ points in the proposed set at each iteration; (ii) \textsc{CHO‑UCB (Random)}, which selects one point uniformly at random from the set; and (iii) \textsc{CHO‑UCB (EHVI)}, (iv) \textsc{CHO‑UCB (Uncertainty)}, and (v) \textsc{CHO‑UCB (SMS‑EGO)}, which select a single point using EHVI, variance, and SMS‑EGO as their secondary selection criteria, respectively.}
To match a total budget of 50 and 100 evaluations, we evaluate in batches of size 
$M$ until the budget is exhausted. In the final iteration, if the remaining evaluation budget is less than $M$, we randomly select the required number of points from the last batch so that the total number of observations matches the prescribed budget.

\begin{figure}[H]
	\centering
 \includegraphics[width=0.95\linewidth]{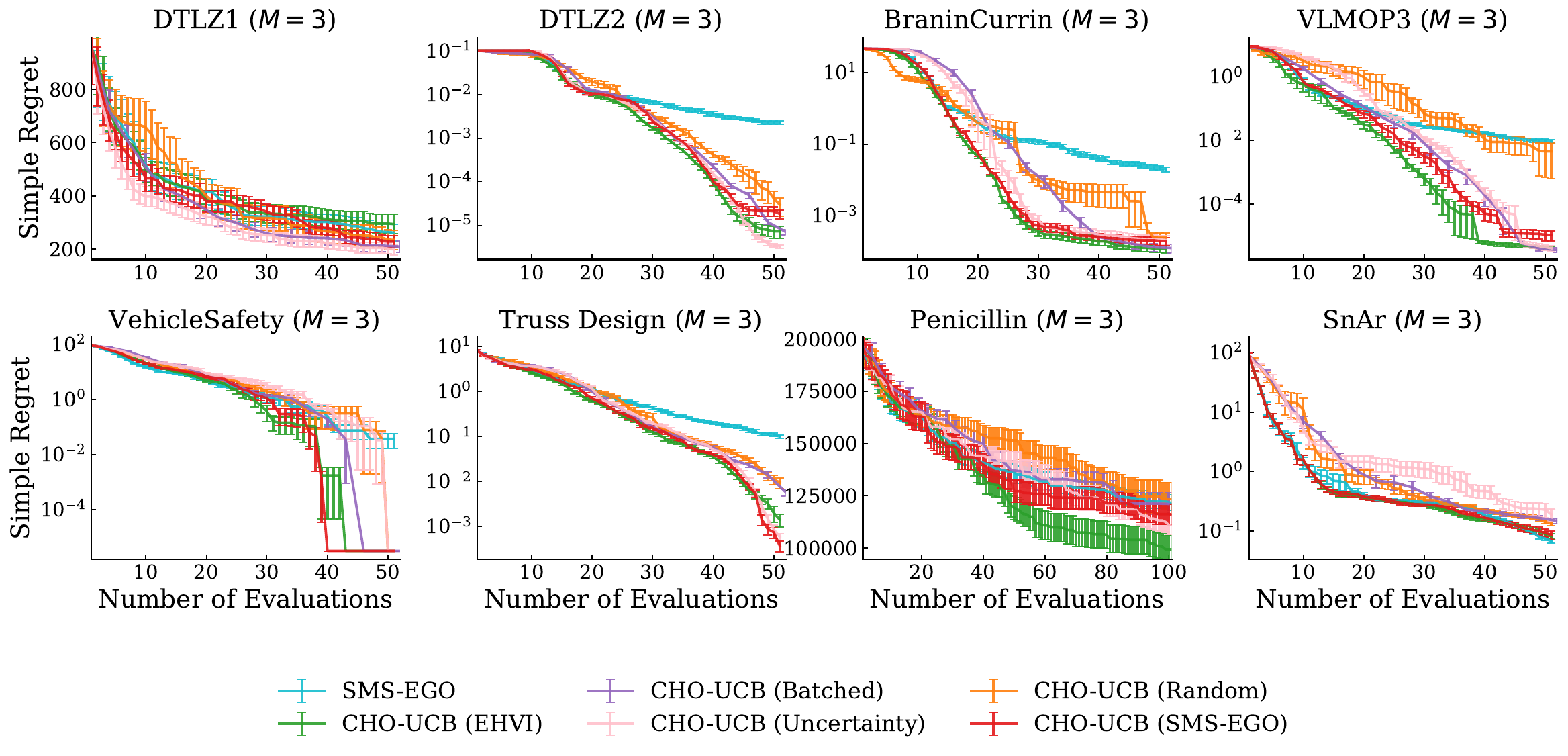}
    \vspace*{-1mm}
	\caption{Scheduler ablation for CHO-UCB with $M=3$. The three schedulers (EHVI, SMS‑EGO, and Uncertainty) perform similarly, indicating that performance is not sensitive to the choice of scheduler.}
	\label{total_UCB_comparison_with_batch}
\end{figure}

\textcolor{black}{Figure~\ref{total_UCB_comparison_with_batch} shows that \textsc{CHO‑UCB (Random)} performs generally worst among the variants, followed by \textsc{CHO‑UCB (Batch)}. The remaining three schedulers (EHVI, SMS‑EGO, and Uncertainty) perform similarly, indicating that performance is not sensitive to the choice of scheduler. This insensitivity confirms that our algorithm's performance stems fundamentally from the cardinality‑aware set construction, rather than from reliance on any specific heuristic.}


\subsection{Study of CHO-EI}
\label{Study of CHO-EI}
In this section, we examine the behavior of CHO-EI.
We first identify that its main challenge arises from the optimization of the acquisition function itself.
Subsequently, we evaluate the MOBO performance of CHO-EI under several acquisition function optimization strategies,
SPI, logEI transformation,
and increased numbers of restarts and initial samples.
While some techniques provide noticeable improvements,
the overall performance of CHO-EI remains inferior to that of REHVI.

\subsubsection{Challenges in Optimizing the CHO-EI Acquisition Function}
\label{acquisition_value}
 The difficulty in CHO-EI's acquisition optimization stems from its combinatorial nature. Although we formulate this as an $M\times d$-dimensional single-objective problem, it differs fundamentally from traditional single-objective BO. While standard BO employs an $M\times d$-dimensional GP as the surrogate model, our approach is based on $K$ $d$-dimensional GPs that are joined to $M\times d$ dimensions through the hypervolume operation.

This combinatorial structure significantly impacts the optimization landscape. For example, in the case of $M=2$, 20 observations, this  means $20^2 = 400$ known data points in the $M\times d$ space due to all possible pairwise combinations. This dense combinatorial coverage generally reduces model uncertainty compared to standard $M\times d$-dimensional BO, making it more challenging to identify regions with positive expected improvement values.

\begin{figure}[H]
	\centering
 \includegraphics[width=0.8\linewidth]{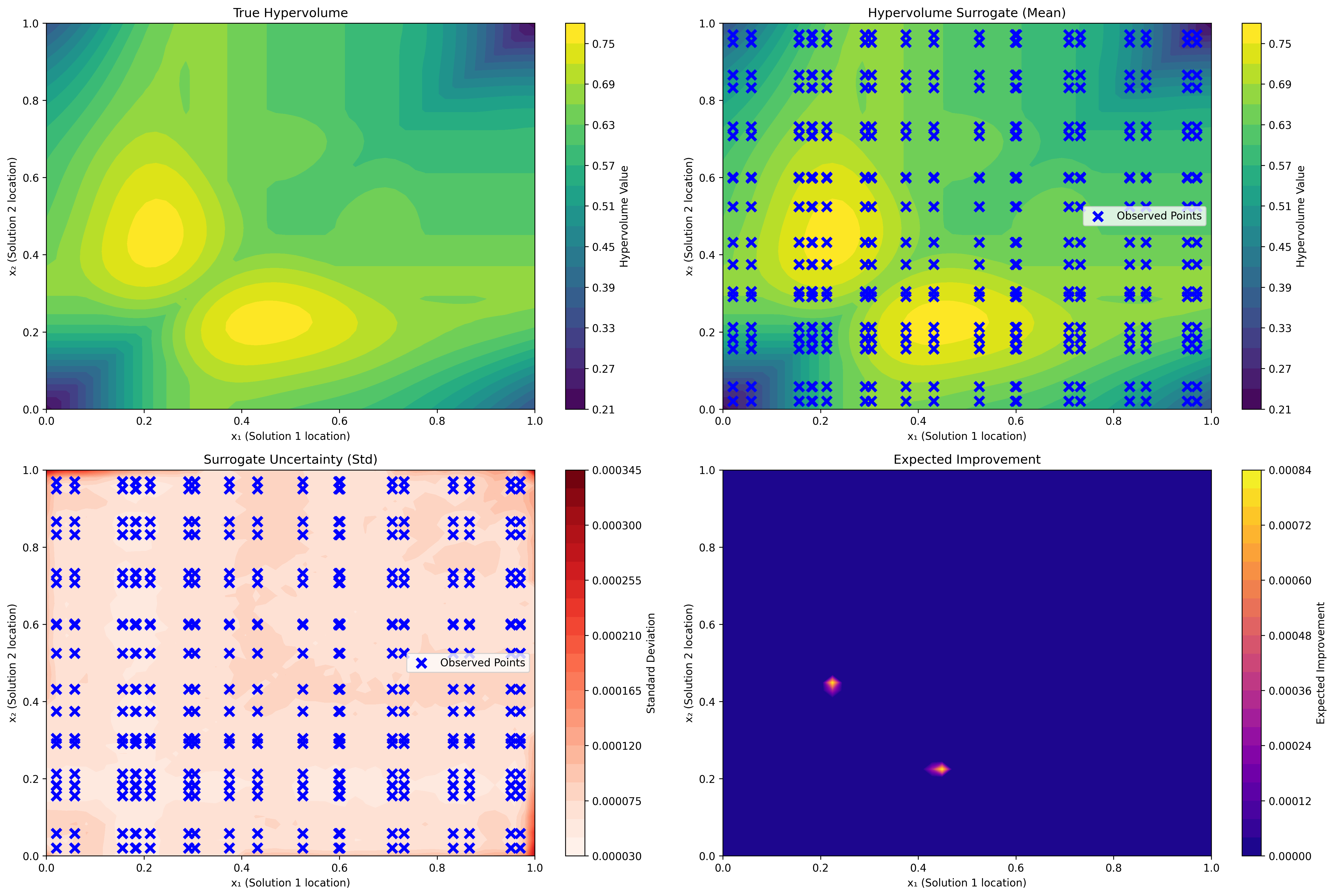}
    \vspace*{-1mm}
	\caption{Combinatorial 2D Surrogate Model Using 1D Gaussian Processes and Hypervolume Operation (20 training data)}
	\label{CHO_EI_explanation_1}
\end{figure}

\begin{figure}[H]
	\centering
 \includegraphics[width=0.8\linewidth]{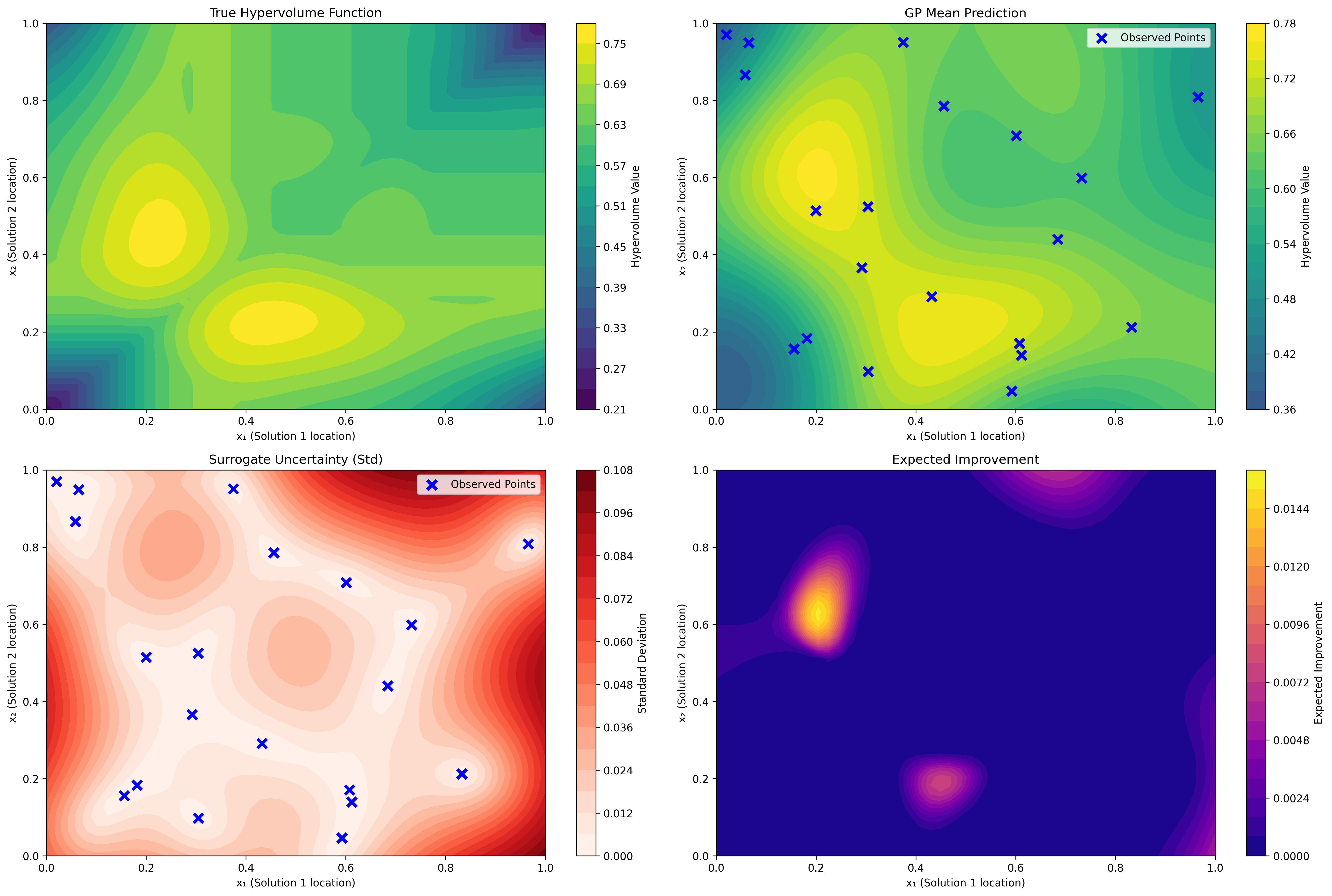}
    \vspace*{-1mm}
	\caption{2D Surrogate Model Directly Modeling Hypervolume Values (20 training data)}
	\label{CHO_EI_explanation_2}
\end{figure}

To illustrate this challenge, we compare two approaches using a 2-objective optimization problem ($f_1(x)=0.8 - (x - 0.3)^2 + 0.1\cdot \text{sin}(10x)$ and $f_2(x)=0.9 - (x - 0.7)^2 - 0.1 \cdot \text{cos}(8x)$) with 20 observations. In Figure~\ref{CHO_EI_explanation_1}, we build separate 1D GP models for each objective and estimate the hypervolume mean and uncertainty through the hypervolume operation (this represents our proposed approach). In Figure~\ref{CHO_EI_explanation_2}, we directly model the hypervolume using a standard 2D GP with 20 observations for comparison.

The results clearly demonstrate the challenge in optimizing the acquisition function. Figure~\ref{CHO_EI_explanation_1} shows that our hypervolume-based approach produces low uncertainty and a nearly flat expected improvement landscape with only small regions of positive values with only $20$ training data. In contrast, Figure~\ref{CHO_EI_explanation_2} shows that the direct 2D GP approach exhibits higher uncertainty and a smoother, more navigable expected improvement surface. This comparison highlights why acquisition optimization becomes significantly more difficult in the hypervolume-based setting.

\subsubsection{SPI}
The experimental results with and without SPI are presented in the Figure~\ref{Ablation_RAASP}. Although SPI generally improves CHO-EI performance, it still remains inferior to REHVI and CHO-UCB.

\subsubsection{CHO-logEI}
\label{CHO-logEI vs CHO-EI}
As discussed earlier, the poor performance of CHO-EI stems from its tendency to exhibit flat regions with zero acquisition values, leading to vanishing gradients. A common strategy to mitigate this issue is logEI transformation \citep{ahmadianshalchi2024pareto}, which preserves gradient information even when the EI value is close to zero. Hence, we conducted an additional experiment to evaluate the effect of applying a logarithmic transformation in the CHO-EI Monte Carlo simulation \citep{ahmadianshalchi2024pareto}, to see whether this trick can solve the issue. Specifically, we use the logmeanexp transformation in \texttt{BoTorch} to compute a log-space Monte Carlo estimate of the expected hypervolume improvement.

\begin{figure}[H]
	\centering
 \includegraphics[width=0.95\linewidth]{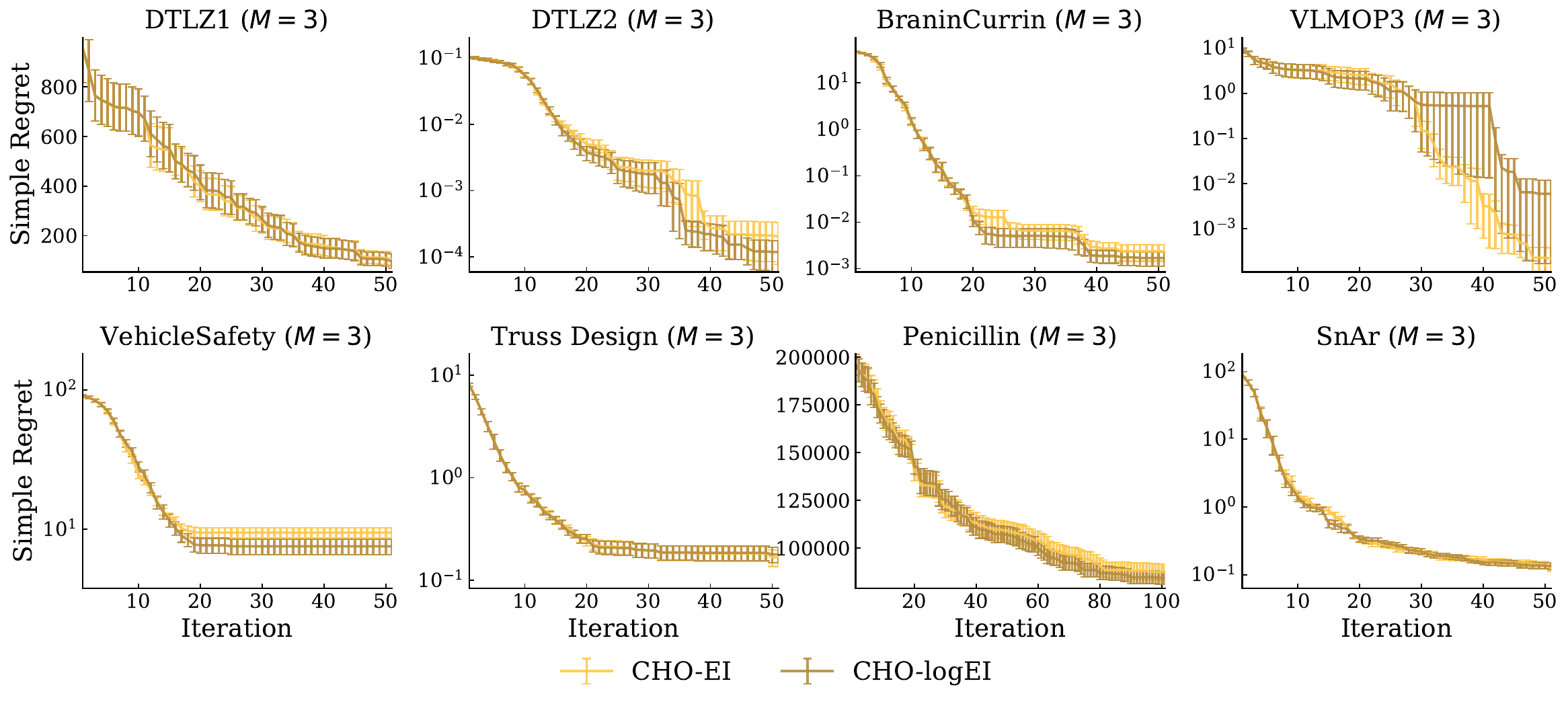}
    \vspace*{-1mm}
	\caption{logEI does not improve performance in the MOBO-CAPS problem.}
	 \label{CHO_logEI}
\end{figure}

We evaluate the case $M=3$, and the results are shown in Figure~\ref{CHO_logEI}. The logarithmic EI transformation does not significantly improve the overall MOBO performance.

\subsection{Different Hyperparameters}
In this section, we conduct tests to examine whether increasing the number of restarts or the number of initial samples can improve CHO-EI's performance. Our results indicate that increasing the number of restarts provides minimal improvement, while increasing the number of initial random samples offers modest benefits in certain cases, such as DTLZ2, BraninCurrin, and the Truss Design problem. However, despite these improvements, CHO-EI continues to underperform relative to REHVI.

\begin{figure}[H]
	\centering
 \includegraphics[width=0.95\linewidth]{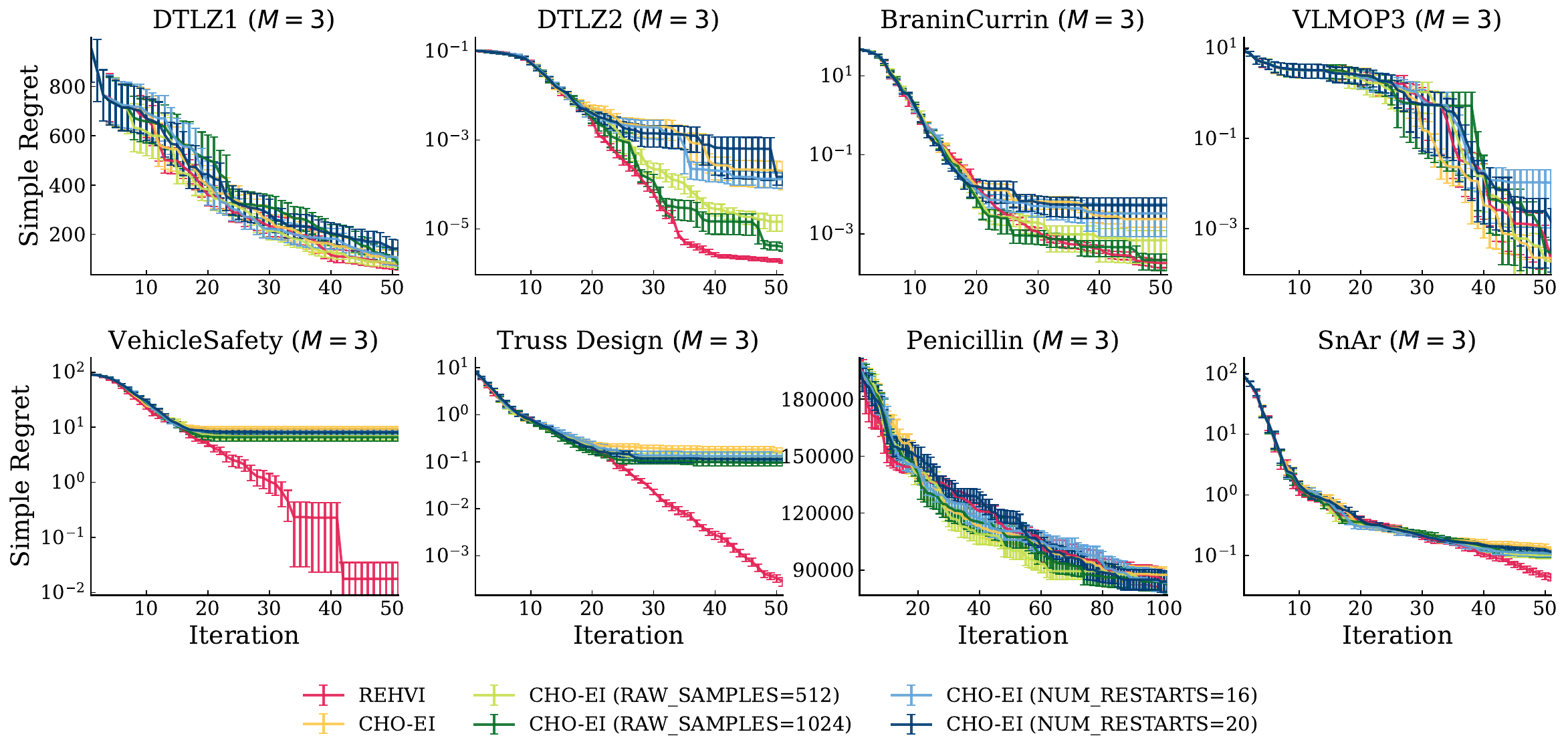}

	\caption{A larger number of initial random samples leads to better MOBO performance; however, REHVI still achieves superior results overall.}
\end{figure}

\newpage
\section{Proof}
\label{full_proof}

\subsection{Distance Metric on Multisets}
\textbf{Populations as multisets.}
We model a population of size $M$ as a collection of $M$ objective vectors in $\mathbb{R}^K$, allowing for duplicates. To formalize this while disregarding the order of elements, we define an equivalence relation $\sim$ on the space of ordered tuples $(\mathbb{R}^K)^M$.
Two tuples $\mathbf{a} = (a^{(1)}, \dots, a^{(M)})$ and $\mathbf{b} = (b^{(1)}, \dots, b^{(M)})$ are considered equivalent, denoted $\mathbf{a} \sim \mathbf{b}$, if one is a permutation of the other. That is, if there exists a permutation $\sigma \in \Pi_M$ such that $b^{(m)} = a^{(\sigma(m))}$ for all $m \in [M]$. The space of multisets is denoted as $\mathcal{M}_M(\mathbb{R}^K)$.\\\\
In this framework, a multiset $X \in \mathcal{M}_M(\mathbb{R}^K)$ represents an equivalence class of tuples that differ only by ordering. Any metric on $\mathcal{M}_M(\mathbb{R}^K)$ must therefore be invariant to the representative sequence chosen.

\begin{definition}[Matching Distance]
\label{matching_distance}
Let $A, B \in \mathcal{M}_M(\mathbb{R}^K)$ be two multisets of size $M$. We define the matching distance $d_{\mathcal{M}}(A, B)$ as the minimum maximal deviation between their elements under any possible bijective pairing. Formally, choosing arbitrary representative sequences $(a^{(1)}, \dots, a^{(M)})$ for $A$ and $(b^{(1)}, \dots, b^{(M)})$ for $B$, we define:
\begin{equation}
    d_{\mathcal{M}}(A, B) ~:=~ \min_{\pi \in \Pi_M} \left( \max_{m \in [M]} \bigl\|a^{(m)} - b^{(\pi(m))}\bigr\|_\infty \right),
\end{equation}
where $\Pi_M$ denotes the set of all permutations of $\{1, \dots, M\}$.
\end{definition}

\begin{proposition}[Metric property of matching distance on multisets]
\label{prop:metric_multiset}
The function $d_{\mathcal{M}}$ is well-defined on
$\mathcal{M}_M(\mathbb{R}^K)$ and is a metric. In particular, for all
$A,B,C\in\mathcal{M}_M(\mathbb{R}^K)$:
\begin{enumerate}
\item \textbf{Nonnegativity:} $d_{\mathcal{M}}(A,B)\ge 0$.
\item \textbf{Identity:} $d_{\mathcal{M}}(A,B)=0 \iff A=B$ as multisets .
\item \textbf{Symmetry:} $d_{\mathcal{M}}(A,B)=d_{\mathcal{M}}(B,A)$.
\item \textbf{Triangle inequality:} $d_{\mathcal{M}}(A,C)\le d_{\mathcal{M}}(A,B)+d_{\mathcal{M}}(B,C)$.
\end{enumerate}
\end{proposition}

\begin{proof}
\textbf{Well-definedness.}
The value of $d_{\mathcal{M}}(X, Y)$ is independent of the specific ordered sequences used to represent the multisets $X$ and $Y$. 
This is immediate from the definition: since the distance is calculated by minimizing over \emph{all possible permutations} $\pi \in \Pi_M$, any initial reordering of the underlying vectors in $X$ or $Y$ does not influence the result. \\\\
\textbf{Nonnegativity.}
For any $\pi$, $\max_m \|a^{(m)}-b^{(\pi(m))}\|_\infty \ge 0$, hence taking the minimum or maximum preserves nonnegativity.\\\\
\textbf{Identity.}
If $A=B$ as multisets, then there exists a permutation $\pi$ such that
$a^{(m)} = b^{(\pi(m))}$ for all $m$.
For that $\pi$, the maximum norm is $0$, hence $d_{\mathcal{M}}(A,B)=0$.
Conversely, if $d_{\mathcal{M}}(A,B)=0$, then for some $\pi$,
$\max_m \|a^{(m)}-b^{(\pi(m))}\|_\infty=0$, so $\|a^{(m)}-b^{(\pi(m))}\|_\infty=0$ for all $m$,
i.e., $a^{(m)}=b^{(\pi(m))}$ for all $m$. Thus $A=B$ as multisets.\\\\
\textbf{Symmetry.}
This follows immediately from the symmetry of the underlying norm and the bijection.
The expression $\min_{\pi} \max_m \|a^{(m)} - b^{(\pi(m))}\|_\infty$ essentially minimizes the maximum distance over all possible one-to-one pairings between the elements of $X$ and $Y$ since the pairing $(x, y)$ has the same distance as $(y, x)$ (i.e., $\|x-y\| = \|y-x\|$), and the set of all possible pairings is identical.\\\\
\textbf{Triangle Inequality:}
Let $A, B, C$ be three multisets. Let $\pi_{AB}$ and $\pi_{BC}$ be the optimal permutations achieving the distances $d_{\mathcal{M}}(A, B)$ and $d_{\mathcal{M}}(B, C)$ respectively.
We construct a candidate permutation $\pi_{AC} = \pi_{BC} \circ \pi_{AB}$ to match $A$ to $C$.
For any index $m$, the triangle inequality of the vector norm implies:
\[
    \| a^{(m)} - c^{(\pi_{AC}(m))} \|_\infty 
    \le 
    \| a^{(m)} - b^{(\pi_{AB}(m))} \|_\infty + \| b^{(\pi_{AB}(m))} - c^{(\pi_{BC}(\pi_{AB}(m)))} \|_\infty.
\]
Taking the maximum over all $m \in [M]$ on both sides:
\begin{align*}
    \max_m \| a^{(m)} - c^{(\pi_{AC}(m))} \|_\infty 
    &\le \max_m \| a^{(m)} - b^{(\pi_{AB}(m))} \|_\infty + \max_m \| b^{(\pi_{AB}(m))} - c^{(\pi_{BC}(\pi_{AB}(m)))} \|_\infty \\
    &= d_{\mathcal{M}}(A, B) + d_{\mathcal{M}}(B, C).
\end{align*}

Finally, since $d_{\mathcal{M}}(A, C)$ minimizes over \emph{all} permutations, we have:
\[
    d_{\mathcal{M}}(A, C) \le \max_m \| a^{(m)} - c^{(\pi_{AC}(m))} \|_\infty \le d_{\mathcal{M}}(A, B) + d_{\mathcal{M}}(B, C).
\]
\end{proof}

\subsection{Lipschitz Continuity of Hypervolume}
\label{sec:lip}

\begin{definition}[Clamped Hypervolume]
\label{def:clamped_hv}
Given a reference point $z_{\mathrm{ref}} \in \mathbb{R}^K$, the hypervolume of a multiset $A \in \mathcal{M}_M(\mathbb{R}^K)$ is defined as the measure of the union of hyper-rectangles dominated by the elements of $A$ and bounded below by $z_{\mathrm{ref}}$. To handle points that do not strictly dominate the reference point, we apply a coordinate-wise projection:
\begin{equation}
    \HV(A) := \lambda^K \left( \bigcup_{a \in A} \prod_{j=1}^K \Bigl[ z_{\mathrm{ref}, j}, \max(a_j, z_{\mathrm{ref}, j}) \Bigr] \right).
\end{equation}
Under this definition, any coordinate $a_j < z_{\mathrm{ref}, j}$ contributes zero length to the volume integration in dimension $j$ (effectively projecting the point onto the reference boundary).
\end{definition}

\begin{assumption}[Bounded objectives]
\label{ass:box}
We fix a reference point $z_{\mathrm{ref}} \in \mathbb{R}^K$ and an upper bound vector $u \in \mathbb{R}^K$ satisfying strict dominance $u \succ z_{\mathrm{ref}}$.
We assume that the true objective functions are bounded from above by $u$:
\[
    f(x) \preceq u \quad \text{for all } x \in \mathcal{X}.
\]
We define the maximum domain width $D_{\max} := \|u - z_{\mathrm{ref}}\|_\infty$. The Lipschitz constant for the hypervolume indicator is derived using this width as $L_{\HV} = M K D_{\max}^{K-1}$.
\end{assumption}

\begin{lemma}[Hypervolume Lipschitzness w.r.t.\ matching distance]
\label{lem:hv-lipschitz}
Under Assumption~\ref{ass:box}, let $A=\!\{a^{(1)},\dots,a^{(M)}\}\!$ and
$B=\!\{b^{(1)},\dots,b^{(M)}\}\!$ be two multisets.
Define
\[
\HV(A) := \lambda^K \left( \bigcup_{a \in A} \prod_{j=1}^K \left[ z_{\mathrm{ref}, j}, \max(a_j, z_{\mathrm{ref}, j}) \right] \right)
\qquad
d_{\mathcal M}(A,B):=\min_{\pi\in\Pi_M}\max_{m\in[M]}\|a^{(m)}-b^{(\pi(m))}\|_\infty.
\]
Then
\[
|\HV(A)-\HV(B)|
\le MKD_{\max}^{K-1}\, d_{\mathcal M}(A,B).
\]
\end{lemma}

\begin{proof}
Let $\pi^\star\in\Pi_M$ attain the minimum in $d_{\mathcal M}(A,B)$ and reindex
$B$ so that $\tilde b^{(m)}:=b^{(\pi^\star(m))}$. Then for all $m$,
\[
\|a^{(m)}-\tilde b^{(m)}\|_\infty \le d_{\mathcal M}(A,B),
\]
and $\HV(B)=\HV(\{\ \tilde b^{(1)},\dots,\tilde b^{(M)}\!\})$ by permutation invariance.

\paragraph{Case 1: All points dominate the reference point.}
Assume first that $a^{(m)} \succeq z_{\mathrm{ref}}$ and $\tilde{b}^{(m)} \succeq z_{\mathrm{ref}}$.
Define intermediate multisets $Z^{(0)},\dots,Z^{(M)}$ by
$Z^{(0)}=A$ and $Z^{(m)}$ is obtained from $Z^{(m-1)}$ by replacing
$a^{(m)}$ with $\tilde b^{(m)}$. By the triangle inequality on $\mathbb{R}$,
\[
|\HV(A)-\HV(B)| \le \sum_{m=1}^M |\HV(Z^{(m-1)})-\HV(Z^{(m)})|.
\]
Let $\{z^{(1)},\dots,z^{(M-1)}\}$ denote the $M-1$ points common to $Z^{(m-1)}$ and $Z^{(m)}$.
Let $U := \bigcup_{\ell=1}^{M-1}\prod_{j=1}^K [z_{\mathrm{ref},j}, z^{(\ell)}_j]$ be the union
of the other $M-1$ boxes, and write
$R:=\prod_{j=1}^K[z_{\mathrm{ref}, j}, a^{(m)}_j]$, $\tilde R:=\prod_{j=1}^K[z_{\mathrm{ref}, j}, \tilde b^{(m)}_j]$.
Since $\HV(Z^{(m-1)})=\lambda^K(U\cup R)$ and $\HV(Z^{(m)})=\lambda^K(U\cup \tilde R)$,
we have
\[
|\HV(Z^{(m-1)})-\HV(Z^{(m)})|
\le \lambda^K\big((U\cup R)\Delta(U\cup \tilde R)\big).
\]
Moreover, $(U\cup R)\Delta(U\cup \tilde R)\subseteq R\Delta \tilde R$.
Consider a point $z \in R \Delta \tilde{R}$. This point belongs to exactly one of the two boxes. Without loss of generality, assume $z \in R$ but $z \notin \tilde{R}$.
Since $z \in R$, we know that for all dimensions $k$, $z_k \le a^{(m)}_k$.
On the other hand, since $z \notin \tilde{R}$, there must be at least one dimension $j$ where the box constraint is violated, meaning $z_j > \tilde{b}^{(m)}_j$.
Combining these inequalities for that specific dimension $j$, the coordinate $z_j$ must lie in the interval $(\tilde{b}^{(m)}_j, a^{(m)}_j]$.
Consequently, $z$ is contained within the slab $S_j$ defined by the mismatch in the $j$-th coordinate:
\[
S_j := \Big\{z\in\mathbb{R}^K:\ z_j\in[\min(a^{(m)}_j,\tilde b^{(m)}_j),\max(a^{(m)}_j,\tilde b^{(m)}_j)],
\ z_\ell\in[z_{\mathrm{ref}, \ell}, u_\ell]\ \forall \ell\neq j\Big\}.
\]
The same logic applies if $z \in \tilde{R} \setminus R$. Therefore,
\[
    R \Delta \tilde{R} \subseteq \bigcup_{j=1}^K S_j.
\]
By subadditivity and Assumption~\ref{ass:box},
\[
\lambda^K(R\Delta \tilde R)
\le \sum_{j=1}^K \lambda^K(S_j)
\le \sum_{j=1}^K |a^{(m)}_j-\tilde b^{(m)}_j|\, D_{\max}^{K-1}
\le K D_{\max}^{K-1}\|a^{(m)}-\tilde b^{(m)}\|_\infty.
\]
Combining the above and summing over $m=1,\dots,M$ yields the result for Case 1.

\paragraph{Case 2: General configuration.}
Consider the general case where elements of $A$ or $B$ may not dominate the reference point $z_{\mathrm{ref}}$ in all coordinates (i.e., $a^{(m)} \not\succeq z_{\mathrm{ref}}$).
We define the projection operator $\operatorname{Proj}_{z_{\mathrm{ref}}} : \mathbb{R}^K \to \mathbb{R}^K$ component-wise by:
\[
    \bigl(\operatorname{Proj}_{z_{\mathrm{ref}}}(x)\bigr)_j = \max(x_j, z_{\mathrm{ref}, j}), \quad \forall j \in \{1, \dots, K\}.
\]
By definition, the hypervolume indicator measures the union of rectangular regions lower-bounded by $z_{\mathrm{ref}}$. Consequently, the contribution of any point $x$ is identical to the contribution of its projection.
Thus, defining the projected multisets $A' = \{\operatorname{Proj}_{z_{\mathrm{ref}}}(a) : a \in A\}$ and $B' = \{\operatorname{Proj}_{z_{\mathrm{ref}}}(b) : b \in B\}$, we have the identity $\HV(A) = \HV(A')$ and $\HV(B) = \HV(B')$.

Since the projected points satisfy $a' \succeq z_{\mathrm{ref}}$ by construction, the condition of Case 1 is met. Therefore, the Lipschitz bound holds for the projected sets:
\[
    |\HV(A) - \HV(B)| = |\HV(A') - \HV(B')| \le MKD_{\max}^{K-1}\, d_{\mathcal M}(A',B').
\]
The operator $\operatorname{Proj}_{z_{\mathrm{ref}}}$ is non-expansive with respect to the $\ell_\infty$-norm, since $|\max(u, c) - \max(v, c)| \le |u - v|$ for any scalars $u, v, c$.
This implies that the matching distance between the projected populations is no larger than the original distance:
\[
    d_{\mathcal M}(A', B') \le d_{\mathcal M}(A, B).
\]
Substituting this back into the inequality yields the final result:
\[
    |\HV(A) - \HV(B)| \le MKD_{\max}^{K-1}\, d_{\mathcal M}(A, B).
\]
\end{proof}

\subsection{Noise-free Simple Regret of CHO-UCB}
\label{sec:cho-ucb-batch-simple}

We consider the noise-free setting where, at each iteration $t=1,\dots,T$,
CHO-UCB selects a population $X_t$ of $M$ points, modeled as a multiset. The proof uses techniques from~\citep{iwazaki2025gaussian, srinivas2009gaussian}.

\paragraph{Setting and notation.}
Let $\mathcal{X} \subset \mathbb{R}^d$ be compact. We have $K$ objectives
$f_i : \mathcal{X} \to \mathbb{R}$, $i \in [K]$, and write
$\mathbf{f}(x) := (f_1(x),\dots,f_K(x)) \in \mathbb{R}^K$. 
In iteration $t$, the algorithm selects a multiset
\[
  X_t = \{x_{t,1},\dots,x_{t,M}\} \in \mathcal{M}_M(\mathcal{X}),
\]
and we observe noiselessly $f(X_t)$.\\\\
We compute hypervolume with respect to a fixed reference point
$z_{\mathrm{ref}}\in\mathbb{R}^K$ (maximization convention), and write
$\HV(\cdot)$ as shorthand for $\HV_{z_{\mathrm{ref}}}(\cdot)$.
We define the optimal CAPS-set of size $M$ as
\[
  X^\star \in \operatorname*{argmax}_{X \in \mathcal{M}_M(\mathcal{X})} \HV\bigl(\mathbf{f}(X)\bigr).
\]
Define the per-iteration regret and simple regret as
\[
  r_t := \HV\bigl(\mathbf{f}(X^\star)\bigr) - \HV\bigl(\mathbf{f}(X_t)\bigr),
  \qquad
  r_T^{\mathrm{simple}}
  := \min_{1 \le t \le T} r_t.
\]

\begin{assumption}[RKHS regularity]
\label{ass:rkhs-batch-simple}
For each $i \in [K]$, $f_i$ lies in the RKHS $\mathcal{H}_{k_i}$ of a known
positive definite kernel $k_i : \mathcal{X} \times \mathcal{X} \to \mathbb{R}$.
Moreover, $k_i(x,x) \le 1$ for all $x \in \mathcal{X}$ and
$\|f_i\|_{\mathcal{H}_{k_i}} \le B$ for some $B > 0$.
\end{assumption}

\begin{lemma}[Objective range under RKHS regularity]
\label{lem:obj-range}
Assume Assumption~\ref{ass:rkhs-batch-simple} and kernel normalization
$k_i(x,x)\le 1$ for all $x\in\mathcal{X}$ and $i=1,\dots,K$.
In the noise-free setting with $\sqrt{\beta_t}=B$, for every $t\ge 1$,
every $x\in\mathcal{X}$, and every objective $i\in\{1,\dots,K\}$,
\[
f_i(x)\in[-B,B],\qquad
\mu_{t-1,i}(x)\in[-B,B],\qquad
\sigma_{t-1,i}(x)\le 1,
\]
and consequently
\begin{equation}
\label{eq:range-ucb-lcb}
\mathrm{UCB}_{t,i}(x),\ \mathrm{LCB}_{t,i}(x)\in[-2B,2B].
\end{equation}
\end{lemma}

\begin{proof}
By the reproducing property, for any $g\in\mathcal{H}_{k_i}$,
$|g(x)|\le \|g\|_{\mathcal{H}_{k_i}}\sqrt{k_i(x,x)}\le \|g\|_{\mathcal{H}_{k_i}}$.
Thus $|f_i(x)|\le \|f_i\|_{\mathcal{H}_{k_i}}\le B$.
In the noise-free setting, the posterior mean $\mu_{t-1,i}$ is the minimum-norm
interpolant (orthogonal projection) of $f_i$ onto
$\mathrm{span}\{k_i(x_s,\cdot)\}_{s\le t-1}$, hence
$\|\mu_{t-1,i}\|_{\mathcal{H}_{k_i}}\le \|f_i\|_{\mathcal{H}_{k_i}}\le B$,
which implies $|\mu_{t-1,i}(x)|\le B$.
Moreover,
$\sigma^2_{t-1,i}(x)=k_i(x,x)-k^\top K^{-1}k\le k_i(x,x)\le 1$,
so $\sigma_{t-1,i}(x)\le 1$.
With $\sqrt{\beta_t}=B$, we have
$\mathrm{UCB}_{t,i}(x)=\mu_{t-1,i}(x)+B\sigma_{t-1,i}(x)\le B+B=2B$
and
$\mathrm{LCB}_{t,i}(x)=\mu_{t-1,i}(x)-B\sigma_{t-1,i}(x)\ge -B-B=-2B$,
giving \eqref{eq:range-ucb-lcb}.
\end{proof}

\paragraph{GP posterior and UCB.}
We place independent zero-mean GP priors on $(f_i)_{i=1}^K$ with kernel
$k_i$. Let $\mu_{t-1,i}$ and $\sigma_{t-1,i}$ denote the posterior mean and
standard deviation of $f_i$ after $(t-1)$ iterations.
In this noise-free RKHS setting, we use a constant exploration parameter:
\begin{equation}
  \label{eq:beta-batch-simple}
  \sqrt{\beta_t} := B
  \quad\text{for all } t \ge 1.
\end{equation}

\paragraph{Algorithm (CHO-UCB).}
In iteration $t = 1,\dots,T$:
\begin{enumerate}
  \item Using the posteriors $(\mu_{t-1,i}, \sigma_{t-1,i})_{i=1}^K$, define
        \[
          UCB_t(x) := \mu_{t-1}(x) + B \sigma_{t-1}(x),
          \qquad
          LCB_t(x) := \mu_{t-1}(x) - B \sigma_{t-1}(x).
        \]
  \item Select a multiset of $M$ points by maximizing the hypervolume of the UCB:
        \[
          X_t \in \operatorname*{argmax}_{X \in \mathcal{M}_M(\mathcal{X})} \HV\bigl(UCB_t(X)\bigr).
        \]
  \item Evaluate $\mathbf{f}(x)$ for all $x \in X_t$ and update all $K$ GP posteriors.
\end{enumerate}

\begin{lemma}[Confidence bound]
\label{lem:conf-batch-simple}
Under Assumption~\ref{ass:rkhs-batch-simple}, for all $i\in[K]$, $t \ge 1$ and
$x \in \mathcal{X}$,
\begin{equation}
  \label{eq:conf-batch-simple}
  |f_i(x) - \mu_{t-1,i}(x)|
  \;\le\;
  B \, \sigma_{t-1,i}(x).
\end{equation}
\end{lemma}

\begin{proof}
This is the standard RKHS inequality for noise-free GP
regression; see Lemma~7 in~\cite{iwazaki2025gaussian}.
\end{proof}

\begin{lemma}[Instantaneous regret bound]
\label{lem:inst-regret-batch}
Under Assumption~\ref{ass:rkhs-batch-simple}, the batched CHO-UCB algorithm satisfies:
\begin{equation}
  \label{eq:r_t-batch}
  r_t
  \;\le\;
  2 B L_{\HV}
  \sum_{i=1}^K \max_{m \in [M]} \sigma_{t-1,i}(x_{t,m}),
\end{equation}
where $L_{\HV} = M K (2B - \min_j z_{\mathrm{ref},j})^{K-1}$ is the Lipschitz constant adapted to the RKHS bounds.
\end{lemma}

\begin{proof}
By Lemma~\ref{lem:conf-batch-simple}, we have for all $x \in \mathcal{X}$ that
$LCB_t(x) \preceq \mathbf{f}(x) \preceq UCB_t(x)$, and thus for any population $X$,
$LCB_t(X) \preceq \mathbf{f}(X) \preceq UCB_t(X)$ componentwise.
Since $\HV(\cdot)$ is monotone w.r.t.\ componentwise dominance, this implies
\[
\HV(\mathbf{f}(X^\star)) \le \HV(UCB_t(X^\star))
\quad\text{and}\quad
\HV(LCB_t(X_t)) \le \HV(\mathbf{f}(X_t)).
\]
Moreover, because $X_t \in \arg\max_{X \in \mathcal{X}^M} \HV(UCB_t(X))$, we have
$\HV(UCB_t(X^\star)) \le \HV(UCB_t(X_t))$.
Combining these inequalities yields
\[
r_t
= \HV(\mathbf{f}(X^\star)) - \HV(\mathbf{f}(X_t))
\le \HV(UCB_t(X_t)) - \HV(LCB_t(X_t)).
\]

From Lemma~\ref{lem:obj-range}, we know that for all $t$ and $x$, every component of $UCB_t(x)$ and $LCB_t(x)$ lies in $[-2B, 2B]$.
Therefore, these sets are contained within the hyper-rectangle bounded above by the vector $\mathbf{u}_{\text{alg}} = (2B, \dots, 2B) \in \mathbb{R}^K$.
Assuming $2B > z_{\mathrm{ref}, j}$ for all $j$, the effective domain width for the Lipschitz constant is:
\[
  D_{\mathrm{max}} := \max_{j \in [K]} (2B - z_{\mathrm{ref}, j}).
\]
We apply Lemma~\ref{lem:hv-lipschitz} using the constant $L_{\HV} = M K D_{\mathrm{max}}^{K-1}$:
\[
  r_t
  \le L_{\HV} \, d_{\mathcal{M}}\bigl(UCB_t(X_t), LCB_t(X_t)\bigr).
\]

Since $UCB_t(X_t)$ and $LCB_t(X_t)$ are computed point-wise on the \emph{same} multiset $X_t$, the optimal pairing is the identity. Thus, the matching distance is bounded by the maximal element-wise difference:
\[
  d_{\mathcal{M}}\bigl(UCB_t(X_t), LCB_t(X_t)\bigr) 
  \le \max_{x \in X_t} \|UCB_t(x) - LCB_t(x)\|_\infty.
\]
By definition, $\|UCB_t(x) - LCB_t(x)\|_\infty = 2B \max_{i} \sigma_{t-1,i}(x)$.
Therefore:
\[
  r_t \le 2 B L_{\HV} \max_{m \in [M]} \max_{i \in [K]} \sigma_{t-1,i}(x_{t,m}).
\]
Using $\max_{i} a_i \le \sum_{i=1}^K a_i$, we obtain the final bound.
\end{proof}


\begin{lemma}[Batched maximum posterior standard deviation bound]
\label{lem:batched-max-sigma}
Fix an objective \(i\in[K]\). Let \(X_t=\{x_{t,1},\ldots,x_{t,M}\}\) be any sequence of batches, and let
\(\sigma_{t-1,i}\) denote the posterior standard deviation after conditioning on all observations in the previous batches \(X_1,\ldots,X_{t-1}\). Then
\[
\sum_{t=1}^T \max_{m\in[M]} \sigma_{t-1,i}(x_{t,m})
\le
C_i(T,d,\theta_i),
\]
where \(C_i(T,d,\theta_i)\) is the cumulative posterior standard-deviation bound in Lemma~\ref{lemma:posterior_std_bound}.
\end{lemma}

\begin{proof}
For each \(t\), choose
\[
x_{t,i}^{\star}\in
\arg\max_{m\in[M]} \sigma_{t-1,i}(x_{t,m}).
\]
Then
\[
\sum_{t=1}^T \max_{m\in[M]} \sigma_{t-1,i}(x_{t,m})
=
\sum_{t=1}^T \sigma_{t-1,i}(x_{t,i}^{\star}).
\]

Now define \(\widetilde{\sigma}_{t-1,i}\) as the posterior standard deviation obtained by conditioning only on the reduced sequential history
\[
\{x_{1,i}^{\star},\ldots,x_{t-1,i}^{\star}\}.
\]
The actual posterior standard deviation \(\sigma_{t-1,i}\) is conditioned on the full previous batched history
\[
\bigcup_{\tau=1}^{t-1}X_\tau,
\]
which contains the reduced history as a subset. Since posterior variance in a noise-free Gaussian process cannot increase when conditioning on additional observations, we have
\[
\sigma_{t-1,i}(x_{t,i}^{\star})
\le
\widetilde{\sigma}_{t-1,i}(x_{t,i}^{\star}).
\]
Therefore,
\[
\sum_{t=1}^T \sigma_{t-1,i}(x_{t,i}^{\star})
\le
\sum_{t=1}^T \widetilde{\sigma}_{t-1,i}(x_{t,i}^{\star}).
\]
Applying Lemma~3 of \citet{iwazaki2025gaussian} to the sequentially updated sequence
\(\{x_{t,i}^{\star}\}_{t=1}^T\) gives
\[
\sum_{t=1}^T \widetilde{\sigma}_{t-1,i}(x_{t,i}^{\star})
\le
C_i(T,d,\theta_i).
\]
Combining the inequalities yields the result.
\end{proof}

\begin{lemma}[cumulative sum of posterior standard deviations for SE and Mat\'ern kernel]
\label{lemma:posterior_std_bound}
Fix any $L > 0$, input domain $\mathcal{X}\subset [-L,L]^d$, and kernel function $k:\mathcal{X}\times \mathcal{X}\to \mathbb{R}$ that satisfies $k(x,x)\leq 1$ for all $x\in \mathcal{X}$. Furthermore, let $C_{\mathrm{SE}},C_{\mathrm{Mat}},\underline{\lambda}_{\mathrm{SE}},\underline{\lambda}_{\mathrm{Mat}}>0,\underline{T}_{\mathrm{SE}},\underline{T}_{\mathrm{Mat}}\geq 2$ be constants such that:
\begin{align*}
\forall \lambda \in (0,\underline{\lambda}_{\mathrm{SE}}],\forall t\geq \underline{T}_{\mathrm{SE}},\quad &\gamma_t(\lambda^2)\leq C_{\mathrm{SE}}\left(\ln \frac{t}{\lambda^2}\right)^{d+1},\\
\forall \lambda \in (0,\underline{\lambda}_{\mathrm{Mat}}],\forall t\geq \underline{T}_{\mathrm{Mat}},\quad &\gamma_t(\lambda^2)\leq C_{\mathrm{Mat}}\left(\frac{t}{\lambda^2}\right)^{\frac{d}{2\nu+d}}\left(\ln \frac{t}{\lambda^2}\right)^{\frac{2\nu}{2\nu+d}}
\end{align*}
for $k=k_{\mathrm{SE}}$ and $k=k_{\mathrm{Mat}\acute{e}rn}$, respectively. Then, the following statements hold for any $T\in \mathbb{N}_+$ and any input sequence $x_1,\ldots,x_T\in \mathcal{X}$.

\textbf{For $k=k_{\mathrm{SE}}$:}
\begin{align*}
&\sum_{t=1}^T \sigma(x_t;\mathbf{X}_{t-1}) \leq \overline{T}_{\mathrm{SE}} + (d+1)\left(\frac{\overline{C}_{\mathrm{SE}}}{2}\right)^{-\frac{3d+3}{2}} \Gamma\left(\frac{3d+3}{2}\right),
\end{align*}
where $\overline{C}_{\mathrm{SE}} = (6C_{\mathrm{SE}})^{-\frac{1}{d+1}}$ and $\overline{T}_{\mathrm{SE}} = \max\{\underline{T}_{\mathrm{SE}},\underline{T}_{\mathrm{SE}}^{(A)},\lfloor(d+1)^{d+1}/\overline{C}_{\mathrm{SE}}^{d+1}\rfloor+1\}$ with $\underline{T}_{\mathrm{SE}}^{(A)} = \min\{T\in \mathbb{N}_+ \mid \forall t\geq T,\ t\exp(-\overline{C}_{\mathrm{SE}}t^{1/(d+1)})\leq \underline{\lambda}_{\mathrm{SE}}^2\}$.

\textbf{For $k=k_{\mathrm{Mat}\acute{e}rn}$ with $\nu>1/2$:}
\begin{align*}
&\sum_{t=1}^T \sigma(x_t;\mathbf{X}_{t-1}) \leq 
\begin{cases}
\overline{T}_{\mathrm{Mat}} + \overline{C}_{\mathrm{Mat}}^{1/2} \dfrac{d}{d-\nu} T^{\frac{d-\nu}{d}} (\ln T)^{\frac{2\nu}{d}} & \text{if } d>\nu, \\
\overline{T}_{\mathrm{Mat}} + \overline{C}_{\mathrm{Mat}}^{1/2} (\ln T)^2 & \text{if } d=\nu, \\
\overline{T}_{\mathrm{Mat}} + \overline{C}_{\mathrm{Mat}}^{1/2} \dfrac{\Gamma\left(\frac{2\nu}{d}+1\right)}{\left(\frac{2\nu}{d}-1\right)^{\frac{2\nu}{d}+1}} & \text{if } d<\nu,
\end{cases}
\end{align*}
where $\overline{C}_{\mathrm{Mat}} = \max\left\{1,\left(2+\frac{2\nu}{d}\right)^{\frac{2\nu}{d}}\left(6C_{\mathrm{Mat}}\right)^{1+\frac{2\nu}{d}}\right\}$ and $\overline{T}_{\mathrm{Mat}} = \max\{4,\underline{T}_{\mathrm{Mat}},\underline{T}_{\mathrm{Mat}}^{(A)}\}$ with $\underline{T}_{\mathrm{Mat}}^{(A)} = \min\{T\in \mathbb{N}_+ \mid \forall t\geq T,\ \overline{C}_{\mathrm{Mat}} t^{-\frac{2\nu}{d}} (\ln t)^{\frac{2\nu}{d}} \leq \underline{\lambda}_{\mathrm{Mat}}^2\}$.
\end{lemma}

\begin{theorem}[Noise-free Simple Regret of Batched CHO-UCB]
Suppose the objective functions $f_i$ satisfy the RKHS Assumption~\ref{ass:rkhs-batch-simple_main} with bound $B$, and the problem domain is bounded as per Assumption~\ref{ass:box_main} with $u = (2B, \dots, 2B)$.
Setting the exploration parameter $\sqrt{\beta_t} = B$ for all $t$, the simple regret of Batched CHO-UCB satisfies:
\[
  r_T^{\mathrm{simple}}
  \;\le\;
  \frac{2 B MK(2B - \min_j z_{\mathrm{ref}, j})^{K-1}}{T} \sum_{i=1}^K C_i(T, d, \theta_i),
\]
where $C_i(T, d, \theta_i)$ represents the cumulative standard deviation bound given by Lemma 3 of \citet{iwazaki2025gaussian}, and $\theta_i$ are the hyperparameters of the $i$-th objective.

\end{theorem}

\begin{proof}
By Lemma~\ref{lem:inst-regret-batch}, the instantaneous regret is bounded by the maximum of standard deviations in the batch.
Summing over $t = 1,\dots,T$:
\[
  \sum_{t=1}^T r_t
  \le
  2 B L_{\HV}
  \sum_{t=1}^T \sum_{i=1}^K \max_{m \in [M]} \sigma_{t-1,i}(x_{t,m}),
\]
where $L_{\HV} = M K (2B - \min_j z_{\mathrm{ref},j})^{K-1}$.

By Lemma~\ref{lem:batched-max-sigma}, we have:
\[
  \sum_{t=1}^T r_t \le 2 B L_{\HV} \sum_{i=1}^K C_i(T,d,\theta_i).
\]
Finally, since $r_T^{\mathrm{simple}} \le \frac{1}{T} \sum_{t=1}^T r_t$, we conclude:
\[
  r_T^{\mathrm{simple}}
  \leq  \frac{2BL_{\HV}}{T} \sum_{i=1}^K C_i(T,d,\theta_i) .
\]
\end{proof}


\newpage
\section*{NeurIPS Paper Checklist}

\begin{enumerate}

\item {\bf Claims}
    \item[] Question: Do the main claims made in the abstract and introduction accurately reflect the paper's contributions and scope?
    \item[] Answer: \answerYes{} 
    \item[] Justification: All contributions and scope are well reflected in the paper’s abstract and introduction.

\item {\bf Limitations}
    \item[] Question: Does the paper discuss the limitations of the work performed by the authors?
    \item[] Answer: \answerYes{} 
    \item[] Justification:  Yes, we acknowledge this limitation and discuss it explicitly in the main paper.

\item {\bf Theory assumptions and proofs}
    \item[] Question: For each theoretical result, does the paper provide the full set of assumptions and a complete (and correct) proof?
    \item[] Answer: \answerYes{} 
    \item[] Justification: All assumptions are stated explicitly in the paper.

    \item {\bf Experimental result reproducibility}
    \item[] Question: Does the paper fully disclose all the information needed to reproduce the main experimental results of the paper to the extent that it affects the main claims and/or conclusions of the paper (regardless of whether the code and data are provided or not)?
    \item[] Answer: \answerYes{} 
    \item[] Justification: All necessary experimental details have been provided, and the code is publicly available to ensure reproducibility.

\item {\bf Open access to data and code}
    \item[] Question: Does the paper provide open access to the data and code, with sufficient instructions to faithfully reproduce the main experimental results, as described in supplemental material?
    \item[] Answer: \answerYes{} 
    \item[] Justification: Yes, the code is publicly available and implements standard Python operations.

\item {\bf Experimental setting/details}
    \item[] Question: Does the paper specify all the training and test details (e.g., data splits, hyperparameters, how they were chosen, type of optimizer) necessary to understand the results?
    \item[] Answer: \answerYes{} 
    \item[] Justification: All necessary details have been provided and the code is available.

\item {\bf Experiment statistical significance}
    \item[] Question: Does the paper report error bars suitably and correctly defined or other appropriate information about the statistical significance of the experiments?
    \item[] Answer: \answerYes{} 
    \item[] Justification:  Error bars are shown to indicate the significance.

\item {\bf Experiments compute resources}
    \item[] Question: For each experiment, does the paper provide sufficient information on the computer resources (type of compute workers, memory, time of execution) needed to reproduce the experiments?
    \item[] Answer: \answerYes{} 
    \item[] Justification:  All necessary details about computer resources have been provided in the Appendix.
    
\item {\bf Code of ethics}
    \item[] Question: Does the research conducted in the paper conform, in every respect, with the NeurIPS Code of Ethics \url{https://neurips.cc/public/EthicsGuidelines}?
    \item[] Answer: \answerYes{} 
    \item[] Justification:  The paper strictly follows the NeurIPS Code of Ethics.

\item {\bf Broader impacts}
    \item[] Question: Does the paper discuss both potential positive societal impacts and negative societal impacts of the work performed?
    \item[] Answer: \answerYes{}  
    \item[] Justification: The paper briefly discusses societal impacts in the Impact Statement. Potential positive impacts include more efficient multi-objective optimization in scientific and engineering applications. Potential negative impacts are application-dependent, as the method could be used to optimize objectives with harmful downstream consequences.
    
\item {\bf Safeguards}
    \item[] Question: Does the paper describe safeguards that have been put in place for responsible release of data or models that have a high risk for misuse (e.g., pre-trained language models, image generators, or scraped datasets)?
    \item[] Answer: \answerNA{} 
    \item[] Justification: The paper poses no such risks.

\item {\bf Licenses for existing assets}
    \item[] Question: Are the creators or original owners of assets (e.g., code, data, models), used in the paper, properly credited and are the license and terms of use explicitly mentioned and properly respected?
    \item[] Answer: \answerYes{} 
    \item[] Justification: This paper does not utilize external data. All prior work and Python packages used have been properly credited through citations.

\item {\bf New assets}
    \item[] Question: Are new assets introduced in the paper well documented and is the documentation provided alongside the assets?
    \item[] Answer: \answerYes{} 
    \item[] Justification: This paper does not introduce new datasets. The code is provided and opensource.

\item {\bf Crowdsourcing and research with human subjects}
    \item[] Question: For crowdsourcing experiments and research with human subjects, does the paper include the full text of instructions given to participants and screenshots, if applicable, as well as details about compensation (if any)? 
    \item[] Answer: \answerNA{} 
    \item[] Justification: The paper does not involve crowdsourcing.

\item {\bf Institutional review board (IRB) approvals or equivalent for research with human subjects}
    \item[] Question: Does the paper describe potential risks incurred by study participants, whether such risks were disclosed to the subjects, and whether Institutional Review Board (IRB) approvals (or an equivalent approval/review based on the requirements of your country or institution) were obtained?
    \item[] Answer: \answerNA{} 
    \item[] Justification: The paper does not involve human subjects, user studies, or study participants, so there are no participant risks to disclose and no IRB approval was required.

\item {\bf Declaration of LLM usage}
    \item[] Question: Does the paper describe the usage of LLMs if it is an important, original, or non-standard component of the core methods in this research? Note that if the LLM is used only for writing, editing, or formatting purposes and does \emph{not} impact the core methodology, scientific rigor, or originality of the research, declaration is not required.
    \item[] Answer: \answerNA{} 
    \item[] Justification: LLM does not impact the core methodology, scientific rigorousness, or originality of the research.

\end{enumerate}

\end{document}